\documentclass{article}

\usepackage{microtype}
\usepackage{graphicx}
\usepackage{subfigure}
\usepackage{booktabs} 

\usepackage{hyperref}

\usepackage[accepted]{icml2024}

\usepackage{amsmath}
\usepackage{amssymb}
\usepackage{mathtools}
\usepackage{amsthm}
\usepackage{bbm}

\usepackage[capitalize,noabbrev]{cleveref}

\theoremstyle{plain}
\newtheorem{theorem}{Theorem}[section]
\newtheorem{proposition}[theorem]{Proposition}

\theoremstyle{definition}
\newtheorem{definition}[theorem]{Definition}

\theoremstyle{remark}

\newcommand{\diag}{\mathrm{diag}}
\newcommand{\sgn}{\mathrm{sgn}}
\DeclareMathOperator*{\argmax}{arg\,max}

\usepackage[export]{adjustbox}
\usepackage{wrapfig}
\usepackage{capt-of}
\usepackage{subcaption}
\usepackage{graphicx}
\usepackage{float}
\usepackage{caption}
\usepackage[textsize=tiny]{todonotes}

\usepackage{xkeyval,xcolor}
\makeatletter
\newlength{\sfp@hseplen}\newlength{\sfp@vseplen}
\define@cmdkey{subfigpos}[sfp@]{pos}[ul]{}
\define@cmdkey{subfigpos}[sfp@]{font}[\scriptsize]{}
\define@cmdkey{subfigpos}[sfp@]{vsep}[2\baselineskip]{\setlength{\sfp@vseplen}{\sfp@vsep}}
\define@cmdkey{subfigpos}[sfp@]{hsep}[10pt]{\setlength{\sfp@hseplen}{\sfp@hsep}}
\newcommand{\subfigimg}[3][,]{%
  \setkeys{Gin,subfigpos}{pos,font,vsep,hsep,#1}
  \setbox1=\hbox{\includegraphics{#3}}
  \ifnum\pdfstrcmp{\sfp@pos}{ul}=0
    \leavevmode\rlap{\usebox1}
    \rlap{\hspace*{\sfp@hsep}\raisebox{\dimexpr\ht1-\sfp@vsep}{\sfp@font{#2}}}
    \phantom{\usebox1}
  \else\ifnum\pdfstrcmp{\sfp@pos}{ur}=0
    \leavevmode\usebox1
    \llap{\raisebox{\dimexpr\ht1-\sfp@vsep}{\sfp@font{#2}}\hspace*{\sfp@hsep}}
  \else\ifnum\pdfstrcmp{\sfp@pos}{lr}=0
    \leavevmode\usebox1
    \llap{\raisebox{\sfp@vsep}{\sfp@font{#2}}\hspace*{\sfp@hsep}}
  \else\ifnum\pdfstrcmp{\sfp@pos}{lr2}=0
    \leavevmode\usebox1
    \llap{\raisebox{8pt}{\sfp@font{#2}}\hspace*{\sfp@hsep}}
  \else\ifnum\pdfstrcmp{\sfp@pos}{uc}=0
    \leavevmode\rlap{\usebox1}
    \rlap{\hspace*{1.2cm}{\raisebox{2.5cm} {\sfp@font{#2}}}}
    \phantom{\usebox1}
  \else
    \leavevmode\rlap{\usebox1}
    \rlap{\hspace*{\sfp@hseplen}\raisebox{\sfp@vsep}{\sfp@font{#2}}}
    \phantom{\usebox1}
  \fi\fi\fi
}

\icmltitlerunning{Activation-Descent Regularization for Input Optimization of ReLU Networks}

\begin{document}

\twocolumn[
\icmltitle{Activation-Descent Regularization for Input Optimization of ReLU Networks}




\begin{icmlauthorlist}
\icmlauthor{Hongzhan Yu}{yyy}
\icmlauthor{Sicun Gao}{yyy}
\end{icmlauthorlist}

\icmlaffiliation{yyy}{Department of Computer Science \& Engineering, University of California San Diego, USA}

\icmlcorrespondingauthor{Hongzhan Yu}{hoy021@ucsd.edu}

\icmlkeywords{Machine Learning, ICML}

\vskip 0.3in
]



\printAffiliationsAndNotice{}  

\begin{abstract}
We present a new approach for input optimization of ReLU networks that explicitly takes into account the effect of changes in activation patterns. We analyze local optimization steps in both the input space and the space of activation patterns to propose methods with superior local descent properties. To accomplish this, we convert the discrete space of activation patterns into differentiable representations and propose regularization terms that improve each descent step. Our experiments demonstrate the effectiveness of the proposed input-optimization methods for improving the state-of-the-art in various areas, such as adversarial learning, generative modeling, and reinforcement learning. 
\end{abstract}

\section{Introduction}

Many problems in deep learning involves optimizing the inputs of neural networks, instead of the model parameters. Examples include finding effective adversarial attacks~\cite{madry2017towards, wong2018provable, ilyas2019adversarial, xu2020adversarial,zhang2022branch}, optimizing latent variables in generative models~\cite{bojanowski2017optimizing, zadeh2019variational}, finding actions that maximizes Q-values in reinforcement learning~\cite{lillicrap2015continuous, fujimoto2018addressing, fujimoto2019off}, and verification of neural networks~\cite{bunel2018unified, cohen2019certified,shi2023formal}. Such input optimization problems typically involve lower-dimensional problems than the standard parameter optimization problem in network training, but the loss landscapes can be more complex with a stronger need for dealing with the combinatorial nature of it~\cite{agrawal2019differentiable, gurumurthy2021joint}.

The standard approach for input optimization is to follow the gradient of the objective function over the input variables. However, this approach can be problematic, especially when dealing with ReLU networks, as the ReLU activation function is discontinuous and non-differentiable, resulting in a gradient of zero for all negative pre-activation values. This means that the steepest descent direction only accounts for descent within the decision boundaries of a specific ReLU activation pattern. In deep neural networks with a large number of activation patterns, the region of inputs for each pattern is typically small. As a result, following the gradient direction can lead to the input jumping over to a different activation pattern with vastly different numerical properties, resulting in significant deviation from the desired objective direction, even if the step sizes are chosen to be very small.

We develop new methods for input optimization that take into account the impact of changes in activation patterns on the output. To accomplish this, we adopt a dual perspective by considering the local optimization steps both from the input space and the space of activation patterns. 
We convert the original discrete space of activation patterns into differentiable representations to obtain descent directions in the activation pattern space that can be continuously tuned. 
This allows us to introduce new regularization terms for local descent in the input space that encourages the input change to be aligned with the descent directions in the activation space. 
We then form the Lagrangian of the original objective with the regularizers and perform primal-dual descent. The overall procedures can thus achieve better local descent properties than gradient descent and also various forms of randomly perturbed gradient methods. 

Through our experiments\footnote{Codes are available at github.com/hoy021/ADR-GD.}, we show that the proposed methods can significantly improve input optimization.
We use benchmarks from several applications, such as optimizing adversarial attacks to neural image classifiers, reconstructing target images with generative models, and deep reinforcement learning with better action selection.

\section{Related Work}

\textbf{Input optimization in deep learning.} Many problems in deep learning can be formulated as input optimization problems.
One canonical application is to construct adversarial attacks \cite{goodfellow2014explaining, madry2017towards, wong2018provable, ilyas2019adversarial, xu2020adversarial}. This centralizes around optimizing an objective that leads to an incorrect label prediction by a classifier, while constraining the perturbed input to be within a bounded region around the original input. Fast Gradient Sign Method (FGSM) \cite{goodfellow2014explaining}, and its multi-step variant Projected Gradient Descent (PGD) \cite{madry2017towards, croce2019sparse}  are the two most popular strategies, both of which utilize the signs of input gradients in constructing adversarial examples. 
However, they both heavily rely on the approach of standard gradient descent applied to non-convex objectives. Generally speaking, many important inverse problems~\cite{bora2017compressed, rick2017one, ongie2020deep} and auxiliary tasks~\cite{pattanaik2017robust, amos2018differentiable, oikarinen2021robust, yang2022neural} require solving input optimization problems.
\cite{bora2017compressed} and \cite{rick2017one} showed that one can use a pre-trained generative model as a prior to solve a variety of problems such as image reconstruction, denoising and inpainting, by optimizing on the latent space of generative models. \cite{amos2018differentiable} proposed to train differential Model Predictive Control (MPC) as a policy class for reinforcement learning, and construct controls by optimizing the cost and dynamics function both parameterized by ReLU networks. \cite{huang2017adversarial} and \cite{ilahi2021challenges} demonstrate the effectiveness of white-box adversarial attacks on neural network policies. 
\cite{yang2024lyapunov} employs adversarial optimization to generate critical samples for improving the learning of neural Lyapunov-like functions~\cite{chang2019neural, taylor2020learning, yu2023sequential}.
Those adversarial attacks are typically generated by optimizing on the input of neural networks with FGSM or PGD. 
Meanwhile, recent works \cite{amos2017input, makkuva2020optimal} propose new network designs which yield convex network output with respect to its inputs. They allow the potentials of employing more sophisticated optimization algorithms in solving input optimization, but usually at the heavy cost of model expressivity and capacity. 

\textbf{Non-convex optimization.} Previous work on searching second-order stationary points can be divided into two categories. Hessian-based approach~\cite{nesterov2006cubic, curtis2017trust} relies on computing the Hession to distinguish between first- and second-order stationary points. \cite{nesterov2006cubic} designed a cubic regularization of Newton method, and analyzed its convergence rate to the second-order stationary points of non-convex objective. Trust region methods \cite{curtis2017trust} can reach comparable performance if the algorithm parameters are carefully chosen.  
These methods typically require expensive computation of the full Hessian inverse, 
which motivates the attempts to accelerate them by using only Hessian-vector products~\cite{agarwal2016finding, carmon2016gradient, carmon2018accelerated}. 
\cite{carmon2016gradient} applied gradient descent as a subroutine to approximate the cubic-regularized Newton step, in which a Hessian-vector product oracle is required. 
\cite{tripuraneni2018stochastic} utilized a stochastic Hessian-vector product in further accelerating the cubic-regularized Newton method. 

Another line of work shows that it is possible to converge to the second-order stationary points without direct use of the Hessian. 
\cite{ge2015escaping} showed that stochastic gradient descent can converge to a second-order stationary point in polynomial time. Levy et al. \cite{levy2016power} improved the convergence rate with gradient normalization. \cite{jin2017escape} and \cite{guo2020perturbed} proposed to apply gradient perturbations when the second-order information indicates the potential existence of nearby saddle points. It can find a second-order stationary point in a comparable time for converging to a first-order stationary point, as long as the non-convex objective satisfies a Hessian Lipschitz property. 

\section{Preliminaries}

We will focus on feed-forward neural networks with ReLU activation, but the methods can be used in the ReLU components in other neural architectures, such as convolution networks and ResNet, and will be shown in the experiments.

An ${l}$-layer ReLU network defines a vector-valued function ${f}: {\mathbb{R}^n \rightarrow \mathbb{R}^m}$, where ${n}$ and ${m}$ are the input and output dimensions. We write the weight matrices across the $l$ layers as $({W_1, W_2, ..., W_{l+1}})$ and the bias vectors as $({b_1, b_2, ..., b_{l+1}})$, where $W_{l+1}$ and $b_{l+1}$ connect the last layer to the outputs. 
The ReLU activation is ${\sigma(x)=\max(x,0)}$, where ${x}$ can be a vector and the max is taken coordinate-wise. Thus, for any input ${x}{\in\mathbb{R}^n}$, the output of the network ${f(x)} {\in\mathbb{R}^m}$ is determined by the following equations:
\begin{align} \label{nndef}
\begin{split}
    h^{(1)} &= W_1\cdot x + b_1,\\
    h^{(i+1)} &= W_{i+1}\cdot \sigma(h^{(i)}) + b_{i+1},\ \ i \in [1, l], \\
    f(x) &= h^{(l+1)} 
\end{split}
\end{align} 
For $i\in [1,l]$, we write $d_i$ for the dimensionality of each $h^{(i)}$, i.e., the number of activation units in each layer $i$. 
With $d_0=n$ and $d_{l+1}=m$, each $W_i$ is of shape $d_{i}\times d_{i-1}$.


Given an arbitrary input $x$ to the network, the output of each activation unit takes either zero or positive values. Activation units with zero value outputs are called \textit{dead neurons}, and the positive ones are the \textit{active neurons}. 
An \textit{activation pattern} of the network on input ${x}$ is the function 
\begin{definition}[Activation Pattern]
Let the network $f$ be as in (\ref{nndef}), then the activation pattern is a function with
\begin{equation}
A_{f}(i,j)=\sgn(\sigma(h^{(i)}_j (x)))    
\end{equation}
on input $x$ at the $j^{th}$ neuron on the $i^{th}$ layer. 
\end{definition}
It is similar to the definition in ~\cite{hanin2019deep} but with explicit indexing that will be important for defining operations on the activation patterns. For each input, the forward evaluation function has a fixed activation function, under which the network is reduced to an affine function for a small region in the input space that shares same activation pattern. 
Such regions are polytope, as they are defined by linear constraints on the activation units. The total number of such regions is the same as the possible activation patterns, which is exponential in the number of neurons. This combinatorially large space marks the difficulty of input optimization of neural networks, which is clearly NP-hard~\cite{knuth1974postscript, van1991handbook}. 

\begin{figure}[t]
  \centering
  \subfigimg[width=0.23\textwidth, pos = uc, font=\small]{(a)}{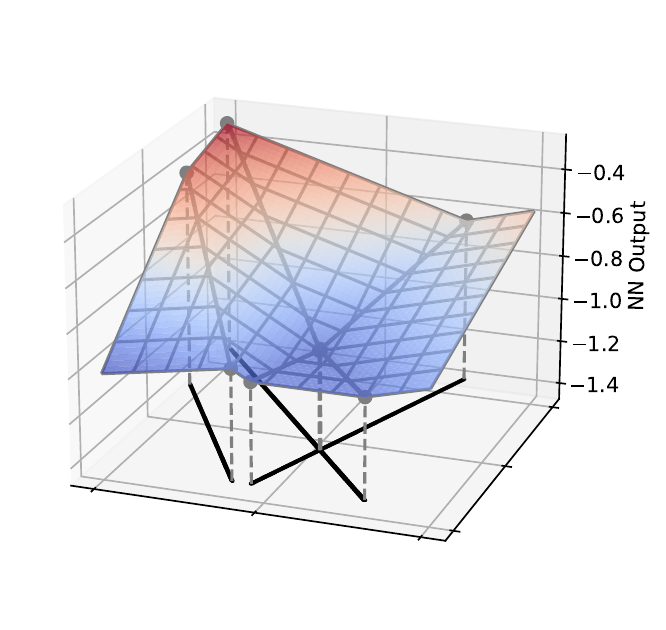}   
  \subfigimg[width=0.23\textwidth, pos = uc, font=\small]{(b)}{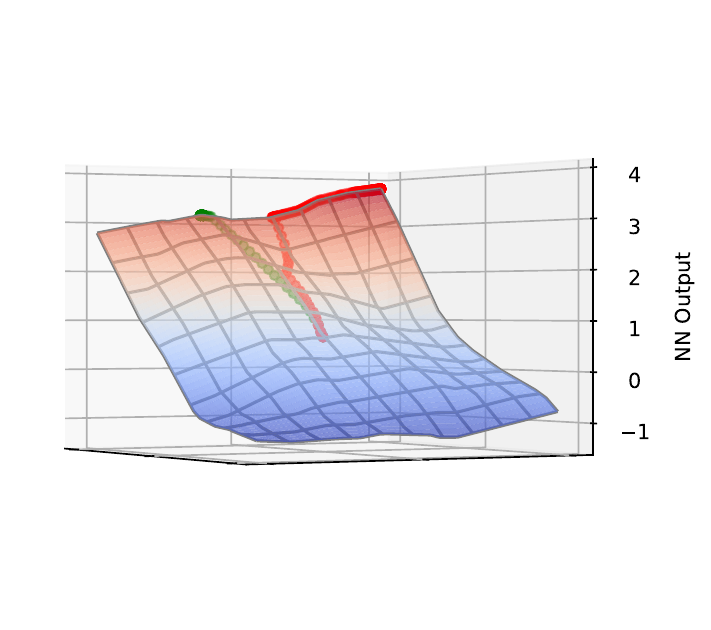}     \\
  \vspace{-1.1cm}
  \subfigimg[width=0.23\textwidth, pos = uc, font=\small]{(c)}{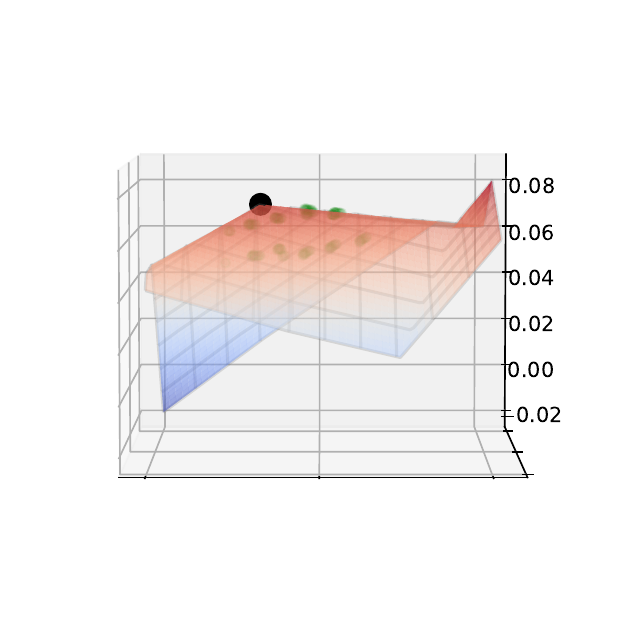}   
  \subfigimg[width=0.23\textwidth, pos = uc, font=\small]{(d)}{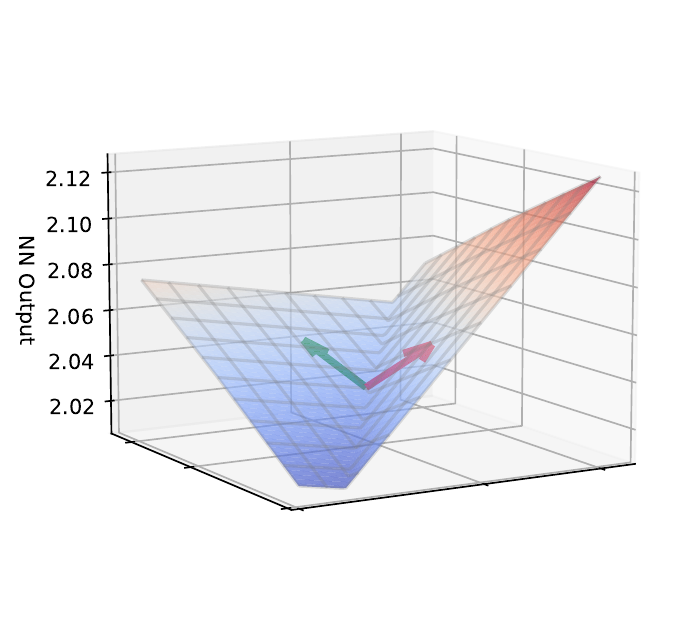}   
  \vspace{-.62cm}
  \caption{\small Value landscape of 2-D ReLU networks with scalar output space. \textbf{(a)} A small ReLU network with 3 hidden neurons. Each solid black line corresponds to one neuron, representing the set of inputs that achieve zero values. \textbf{(b)} The green and red lines demonstrate the trajectories under the optimization of vanilla GD and the proposed algorithm respectively. \textbf{(c)} The scenario where GD optimization gets stuck around a local maxima. The green dots demonstrate the GD steps, and the black dot is the local maxima. \textbf{(d)} The scenario where the input achieves zero value at one neuron. The green and red arrows are ${\nabla_{x}^{-} f(x)}$ and ${\nabla_{x}^{+} f(x)}$ respectively.}
  \label{fig::low_dim_examples}
\end{figure}
\section{Problems with the Input Gradients}
\label{section::gd_problems}

Since the input space is divided by the activation patterns into exponentially many polytope regions, the gradient of the network over the inputs $\nabla_x f(x)$ is determined by an affine function that is only valid for the small region around $x$. 
For illustrative purposes, Figure~\ref{fig::low_dim_examples} (a) visualizes the activation patterns and the corresponding value landscape of a ReLU network with 3 hidden neurons and a scalar field as the output space. The input space is partitioned into 5 polyhedrons, each of which associates with one activation pattern. The network value ${f}$ is convex with respect to the input for the inputs within one polyhedron.
Consequently, when performing input optimization by following the input gradient direction, 
gradient information can produce false optimality analysis for the update steps that lead out of the polytope region corresponding to the activation pattern of $x$.

We re-illustrate the above insights with mathematical formulas.
For an arbitrary input ${x}$, denote the input gradient as ${\nabla_{x} f(x)}$. Define ${\alpha^{*}}$ to be the distance from ${x}$ to the closet polyhedron boundary along ${\nabla_{x} f(x)}$,
that is the optimal step size for descending along ${\nabla_{x} f(x)}$ with improvement guarantee on the target objective.
For all the step sizes ${\alpha}$ that is no larger than ${\alpha^*}$, it should be the case that $f(x + \alpha^* \nabla_{x} f(x)) \geq f(x + \alpha \nabla_{x} f(x)) \geq f(x).$ However, as discussed above, ${\nabla_{x} f(x)}$ is not a reliable descent direction if the activation pattern of new input, i.e. $\tilde{x} = x + \alpha \nabla_{x} f(x)$, is different from that of $x$,
in which case  $\alpha$ must have a value strictly larger than $\alpha^{*}$.
Figure~\ref{fig::low_dim_examples}(c) demonstrates a scenario where applying standard gradient descent to perform input optimization converges to a local maxima.
This is because the selected step size value is not large enough to escape the polytope regions that connect the local maxima.
Increasing the value of step size is never a robust solution, which motivates the development of new techniques to derive better decent directions that look beyond the local regions.

Next, we discuss another typical failure scenario caused by the locality of standard gradient directions.  
On condition that the input variable attains zero value at one hidden neuron, that is when the input falls upon the polyhedron boundary, 
vanilla gradient descent rarely consider the decent directions which turn on that specific neuron.
Figure~\ref{fig::low_dim_examples}(d) provides a scenario where a worse descent direction is selected due to the above issue even though there exists a clearly better gradient direction that leads to more improvement to the objective. 
Denote the gradient directions corresponding to having dead and alive neuron for the intersected hyperplane with ${\nabla_{x}^{-} f(x)}$ and ${\nabla_{x}^{+} f(x)}$ respectively.
The worse direction is selected by following input gradients if the new input led by ${\nabla_{x}^{-} f(x)}$, e.g. $f(x + \alpha\nabla_{x}^{-} f(x)))$, attains a lower objective value than that from $\nabla_{x}^{-} f(x)$ for a small-valued step size $\alpha$.
The exceptional case is when ${\nabla_{x}^{-} f(x)}$ 
induces a negative dot product with the normal vector of the hyperplane for that neuron.
When optimizing for large networks, there can be an extremely large number of hyperplanes, each corresponding to one neuron, accounting for the finite input space. 
In that case, every step of gradient descent optimization can lead the input variable to be near the boundaries for many neurons.
Therefore, this issue occurs frequently when optimizing large networks.

\section{Activation-Descent Regularization}

\subsection{Augmenting Inputs with Activation Variables}
\label{section::definition}

From the analysis above, we see that the main problem of standard input gradients is that they cannot predict how the activation patterns change even locally.
Ideally, we would like to understand how the output of the network changes with respect to the change in the activation patterns, so that the optimization searches beyond the local polyhedron.  
Thus, the first step is to introduce new binary variables ${\nu}$ that can directly represent the activation patterns:
\begin{align*}
    \nu &= \{\nu^{(1)}, ..., \nu^{(l)}\} \in \{0, 1\}^{\text{\#neurons}}, \\  
    \nu^{(i)} &= \{\nu^{(i)}_{1}, ..., \nu^{(i)}_{d_i}\} \in \{0, 1\}^{d_{i}}, \ \ i \in [1, l].\
\end{align*}
Consider a rewriting of the network definition in Eq. (\ref{nndef}) in the following form: 
\begin{align}\label{nndef2}
\begin{split}
    h^{(1)}(x,\nu) &= W_1\cdot x + b_1,\\
    h^{(i+1)}(x,\nu) &= W_{i+1}\cdot \diag(\nu^{(i)})\cdot h^{(i)}(x,\nu) + b_{i+1},\\
    &\hspace{4.3cm}\ \ i \in [1, l], \\
    \hat f(x,\nu) &= h^{(l+1)}(x,\nu), 
\end{split}
\end{align}
where ${\diag(\nu^{(i)}) \in \mathbb{R}^{d_{i} \times d_{i}}}$ is the diagonal matrix of new variables ${\nu^{(i)}}$. 
It is clear that the function defined in (\ref{nndef2}) attains the same value as the neural network in Eq. (\ref{nndef}) if ${\nu}$ variables satisfies:
\begin{eqnarray}\label{feasible}
\nu^{(i)}_{j}=\begin{cases}
    1, & \text{if } h^{(i)}_j(x) \geq 0,\\
    0, & \text{otherwise.}
\end{cases}
\end{eqnarray}
This is because when ${h^{(i)}_{j}(x)\geq 0}$, 
\[\sigma(h^{(i)}_{j}(x))=\nu^{(i)}_{j}h^{(i)}_{j}(x).\]
In fact, the ${h^{(i)}}$ vectors will have exactly the same value as the original network. 

\begin{proposition}
If $x$ and $\nu$ satisfy constraints in (\ref{feasible}), then $f(x)=\hat{f}(x,\nu)$. 
\end{proposition}
Now the newly parameterized function ${\hat{f}(x,\nu)}$ allows us to explicitly inspect how the changes in the activation patterns, captured by the ${\nu}$ variables, affect the changes in the output of the network. In fact we can derive the steepest descent direction in the discrete space of  $\nu$ variables.
This information can be obtained by the partial derivative of the output component ${h^{(l+1)}_k(x,\nu)}$ over the activation variable ${\nu^{(i)}_j}$, which assumes for the time-being that they take continuous values and can be differentiated. 
Then from the sign of the partial derivative, we need to check if ${\nu^{(i)}_{j}}$ can change its actual discrete value accordingly
and stay within $\{0,1\}$. 
In other words, we project the gradient of the output component to the discrete domain of ${\nu}$. Importantly, the projected direction does not need to be a feasible direction for the network, because we are treating ${\nu}$ as independent variables and have not taken into account of the feasibility constraints in Eq. (\ref{feasible}), which will be handled later. 

\begin{definition}[Steepest Activation-Descent Direction]\label{steepest}
Let ${h^{(l+1)}_k}$ be an arbitrary component of the output of the network ${\hat{f}(x,v)}$ defined in Eq (\ref{nndef2}). 
We define the projected descent direction of 
${h^{(l+1)}_k(x, \nu)}$ on ${\nu^{(i)}_j}$, for non-zero ${h^{(i)}_j(x, \nu)}$, as follows:
\begin{align}
\begin{split}
    \partial \nu^{(i)}_j =&  \hat{\sgn}(\nu^{(i)}_j - 0.5) \cdot \min\Big(\displaystyle\hat{\sgn}\bigg(\frac{\partial h^{(l+1)}_k(x,\nu)}{\partial \nu^{(i)}_{j}}\cdot \\ 
    &\frac{1}{h^{(i)}_j(x,\nu)} \bigg) \cdot \hat{\sgn}(\nu^{(i)}_j - 0.5),0\Big)
\end{split}
\label{eqn::steepest_direction}
\end{align}
where ${\hat{\sgn}(x) = 2 \cdot \sgn(x) - 1}$ applies sign operation into the set $\{-1, 1\}$.
For the case ${h^{(i)}_j(x,\nu)} = 0$, we set ${\partial \nu^{(i)}_j=0}$. 
Then the overall steepest descent direction in the discrete domain for $\nu$ is simply the vector consisting of all these components $\partial \nu = [\partial\nu^{(1)}_1, \cdots, \partial\nu^{(l)}_{d_l}]^T$. This direction is the steepest descent direction in the space of discrete values for $\nu$ because it is turning each component on or off based on whether they contribute positively or negatively to the output value.  
\end{definition}

However, the steepest activation-descent direction defined for ${\hat{f}(x,\nu)}$ is only valid if we consider ${\nu}$ as independent from ${x}$ and ignore the constraints in Eq. (\ref{feasible}). Indeed, it is easy to come up with networks that require large changes in ${x}$ for the activation patterns ${\nu}$ to change, which will invalidate the local analysis based on gradients. Consequently, we want to capture the activation-descent direction in a continuous representation of the activation patterns. 
A natural idea is then to replace the discrete-valued activation variables ${\nu}$ by continuous activation functions such as the sigmoids, so that the activation patterns can be tuned continuously. 
However, if we make that change from the discrete values to the continuous ones, the overall output of the network will change in complex ways, because of the dependency across layers. 
It is thus crucial to define surrogate functions such that, even if the value of the surrogate function is different from the original function, the gradient over the new activation variables is consistent with the steepest direction in Definition~\ref{steepest}.
We achieve this by showing that the gradient of the surrogate function over the activation variables always forms a positive inner-product with the steepest direction in the discrete form (see Appendix \ref{appendix::proof}). 

\begin{definition}[Sigmoid-Surrogate Network]
Let ${f(x)}$ be a ReLU network defined in Eq. (\ref{nndef}), and ${\eta} \in [0, 1]^{\text{\# neurons}}$ be the continuous representation of activation patterns. Using the same weight and bias parameters, we define its sigmoid-surrogate network as:
\begin{align}
\begin{split}
    \bar{h}^{(1)}(x, \eta) &= W_{1}\cdot x + b_{1}  \\
    \bar{h}^{(i+1)}(x, \eta) &=W_{i+1}\cdot \diag(s_{\alpha}(\eta^{(i)}))\cdot \bar{h}^{(i)}(x, \eta) \\ 
    &\hspace{0.5cm}+ b_{i+1}, \hspace{2.1cm}\ \ i \in [1, l], \\
    \bar{f}(x, \eta) &= \bar{h}^{(l+1)}(x, \eta)
\end{split}
\label{eq::sigmoid_network}
\end{align}
where $s_{\alpha}(x) = (1 + \exp(-\alpha (x-\frac{1}{2})))^{-1}$ is the sigmoid function with an offset of ${1/2}$.
\end{definition}
The continuous ${\eta}$ variables should satisfy the corresponding feasibility constraints:
\begin{eqnarray}\label{feasible2}
\eta^{(i)}_{j}: \begin{cases}
    \geq \frac{1}{2}, & \text{if } h^{(i)}_j(x) \geq 0,\\
    < \frac{1}{2}, & \text{otherwise.}
\end{cases}
\end{eqnarray}
\begin{proposition}
\label{prop::eta_gradient}
For any ${x}$ and ${\eta}$ that satisfy the feasibility constraints in Eq.~(\ref{feasible2}), the gradient on ${\eta}$ defines descent directions of the original function.  
\end{proposition}

\subsection{Optimization Objective}

Next, we discuss the objective for optimizing ${x}$ and ${\eta}$ variables. 
Let $[x]_{+} = \max(x, 0)$ for $x \in \mathbb{R}$, to differentiate itself from the activation function $\sigma$.
Given a $l$-layer ReLU network and an objective function ${J}$ to \textit{maximize}, we perform optimization to minimize the following objective ${L^*}$: 
{\small\begin{align}
    L_{i}(x, \eta) &= \sum_{j = 1}^{d_{i}} \bigg(\Bigg[h^{(i)}_j(x) / \|P^{(i)}_{j}\| \cdot \Big[\frac{1}{2} - \eta^{(i)}_{j}\Big]_{+}\Bigg]_{+}\nonumber\\
    &\hspace{-0.5cm}+\Bigg[- h^{(i)}_j(x) / \|P^{(i)}_{j}\| \cdot \Big[\eta^{(i)}_{j} - \frac{1}{2}\Big]_{+}\Bigg]_{+}\bigg), \label{eq::constraint_loss}\\ 
    L^{*}(x, \eta) &= -J(\bar{h}^{(l+1)}(x, \eta)) + \beta \sum_{i=1}^{l} L_{i}(x, \eta). \label{eq::objective}  
\end{align}}
where ${\beta}$ is a scalar parameter, 
and ${P^{(i)}_{j}}$ is the $j^{th}$ row vector of matrix ${P^{(i)} \in \mathbb{R}^{d_i \times n}}$ representing the hyperplane normal vectors for the neurons in the $i^{th}$ layer of the sigmoid-surrogate network:
{\small\begin{align}
    P^{(i)} = W_{i} \cdot \prod_{j=-(i-1)}^{-1} (\diag(s_{\alpha}(\eta^{(-j)})) \cdot W_{-j}).
\end{align}}The first term in (\ref{eq::objective}) is the objective function applied to the output of sigmoid-surrogate network.
The second term penalizes the misclassifications from ${\eta}$ with respect to the ground-truth activation patterns.
It is essential to apply normalization based on ${P}$, as the magnitude order of hidden neurons in ReLU networks varies drastically from layer to layer.
Without normalization, constraint loss (\ref{eq::constraint_loss}) from deeper layers dominates the second term in (\ref{eq::objective}).

\begin{figure}[t]
  \centering
    \subfigure{           
        \includegraphics[width=0.92\linewidth,]{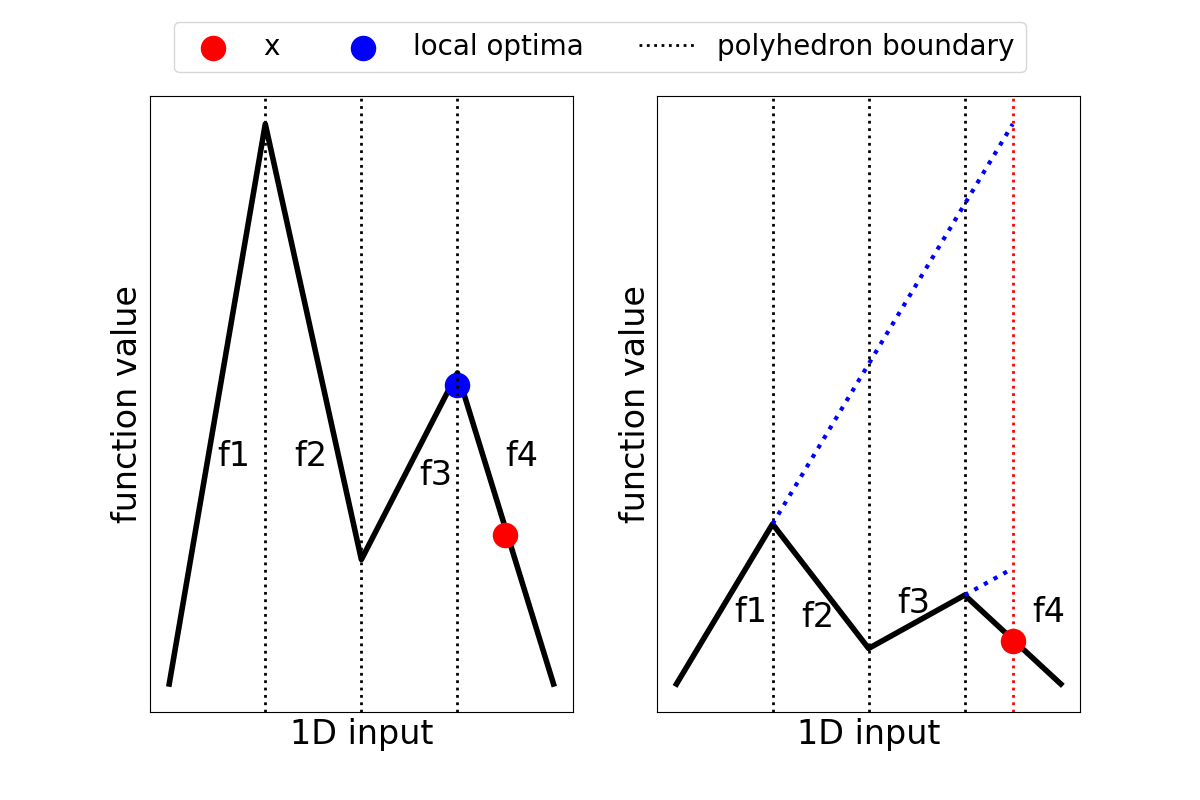} 
    }  
    \vspace{-0.4cm}
  \caption{\small Toy example to illustrate the intuition behind optimizing activation pattern variables $\eta$.
    We consider to maximize a $1$-dimensional piecewise-linear function where each linear segments (e.g. $f_1$-$f_4$ functions) corresponds to one activation pattern.
    \textbf{Left}: Applying vanilla GD at $x$ quickly gets stuck at the nearby local optima.
    \textbf{Right}: Optimizing (\ref{eq::objective}) over $\eta$ identifies promising changes to $\eta$   that \textit{locally} optimizes the overall objective.
    In the above example, the activation pattern underneath $f_1$ will be targeted, as $f_{1}(x) > f_{3}(x) > f_{4}(x)$, despite the inconsistency between the target activation pattern and the current input.
    Optimizing (\ref{eq::constraint_loss}) helps to correct such inconsistencies.
  }
  \label{fig:toy_example}
\end{figure}

Next, we discuss the gradient of (\ref{eq::objective}) on ${x}$ in (\ref{eq::input_gradient}).
The first term in (\ref{eq::input_gradient}) corresponds to the first term in (\ref{eq::objective}).
It approximates the input gradient from standard GD, i.e. 
${\partial J(f(x))/\partial x}$, 
if the feasibility constraints in Eq.~\ref{feasible2} are satisfied. 
In words, this term performs local search for better input values within the local polyhedron as vanilla GD.
The second term in (\ref{eq::input_gradient}) corresponds to the constraint loss (\ref{eq::constraint_loss}) in ${L^{*}}$, optimizing ${x}$ towards the input polyhedron recognized by ${\eta}$.
More specifically, the binary indicator functions (i.e. the $\sgn$ functions) check the inconsistency between what ${\eta}$ captures and the activation patterns of the original network, i.e. ${f(x)}$ instead of ${\bar{f}(x, \eta)}$.
For one specific neuron, at most one of the two indicators can be triggered, which indicates the inconsistency at that neuron.
In that case, the gradient of (\ref{eq::constraint_loss}) optimizes ${x}$ along the hyperplane normal vector of that particular neuron until the input variable reaches the boundary and thus flips the activation pattern.
Intuitively, this term is important in searching more global descent directions that lead ${x}$ to the potentially better polyhedron regions based on the optimization of ${\eta}$.
{\small\begin{align}
    \nabla_{x} L^{*} &= 
    \frac{\partial J(\bar{h}^{(l+1)}(x, \eta))}{\partial \bar{h}^{(l+1)}(x, \eta)} \cdot P^{(l+1)} + \beta \sum_{i=1}^{l} \sum_{j=1}^{d_i}\bigg[ \nonumber\\
    &\hspace{-0.29cm}\sgn\Big(h^{(i)}_{j}(x) \cdot \Big[\frac{1}{2} - \eta^{(i)}_{j}\Big]_{+}\Big) \cdot \frac{\partial h^{(i)}_{j}(x, \eta)}{\partial x} /  \|P^{(i)}_{j}\| \label{eq::input_gradient} \\
    &\hspace{-0.7cm} - \sgn\Big(- h^{(i)}_{j}(x) \cdot \Big[\eta^{(i)}_{j} - \frac{1}{2}\Big]_{+}\Big) \cdot \frac{\partial h^{(i)}_{j}(x, \eta)}{\partial x} /  \|P^{(i)}_{j}\| \nonumber \bigg] \nonumber
\end{align}}
Lastly, we discuss the gradient of constraint loss (\ref{eq::constraint_loss}) on ${\eta}$. 
Without loss of generality, we focus on one arbitrary pattern variable ${\eta^{(i)}_{j}}$:
{\small\begin{align*}
    \frac{\partial L_{i}}{\partial \eta^{(i)}_{j}} &= \displaystyle\hat{\sgn}\Big(h^{(i)}_{j}(x)\Big) \cdot \sgn\Big(h^{(i)}_{j}(x) \cdot (\frac{1}{2} - \eta^{(i)}_{j})\Big) \cdot \frac{h^{(i)}_{j}(x)}{\|P^{(i)}_{j}\|}. 
\end{align*}} This term optimizes ${\eta}$ to satisfy the feasibility constraints in (\ref{feasible2}) with respect to ${x}$. 
If there is detected inconsistency between ${\eta}$ and  the ground-truth activation pattern of ${x}$, this gradient optimizes ${\eta}$ to match the correct activation pattern. 
The division with $\|P^{(i)}_{j}\|$ applies normalization to derive the normalized distance from ${x}$ to the hyperplane of one arbitrary neuron.
We emphasize that the gradients of the first term in (\ref{eq::objective}) on ${\eta}$ are not discarded. 
As these gradients capture the descent directions of the objective on $\eta$ (Proposition \ref{prop::eta_gradient}), they can search for the promising changes to the activation patterns that locally optimize the overall objective despite introducing mismatch between the current input and the target pattern assignments (Figure \ref{fig:toy_example}).
Optimizing the feasibility constraint (\ref{eq::constraint_loss}) helps to enforce consistency between the input variable and the targeted activation patterns.

\subsection{Overall Algorithm}
\label{section::algorithm}

We provide the complete procedures in Algorithm \ref{alg::proposed}, in which 
Perturbed Gradient Descent (Perturbed GD)~\cite{jin2017escape} is employed over ${\eta}$ variables. At line 2, we initialize ${\eta}$ to match the activation patterns of the initial input variable. 
Perturbations are applied to ${\eta}$ only when the gradient norm of $L^{*}$ on ${\eta}$ is lower than the threshold ${\delta}$ (line 8).
Intuitively, randomized perturbations are not helpful unless the variables are currently close to some potential saddle points.
We apply perturbations at most once every \textbf{$T_{p}$} iterations. 
We do not apply perturbations to ${x}$ as the input gradient of the first term in (\ref{eq::objective}) is piecewise and discontinuous. Therefore, if Perturbed GD is deployed for our problem, it applies perturbations as long as the objective landscape is flat enough within the current input polyhedron, regardless of the input being around saddle points or not.
At lines 15-18, we adjust the coefficient parameter ${\beta}$ based on its gradient of ${L^{*}}$ which essentially is the non-negative constraint loss (\ref{eq::constraint_loss}). If the constraint loss has a trivial value, i.e. its norm being less than the tolerance ${\delta_{\beta}}$, we linearly decay ${\beta}$ by ${\gamma}$ to mitigate the weighing of constraint loss.

\section{Experiments}

\subsection{Adversarial Optimization}
\label{section::adversarial_attacking}

\begin{algorithm}[tb]
\caption{Activation-Descent Regularization GD (ADR-GD)}
\label{alg::proposed}
\begin{algorithmic}[1]
\STATE\textbf{Input} $l$-layer ReLU network $f$, total iterations $T$, initial coefficient $\beta_0$, step sizes ($\alpha_{x}$, $\alpha_{\eta}$, $\alpha_{\beta}$), perturbation scale $r$, perturbation frequency $T_{p}$, coefficient decay rate $\gamma$, gradient tolerances ($\delta$, $\delta_{\beta}$) 
\STATE \textbf{Initialize} $x$ and $\eta^{(1)}, ..., \eta^{(l)}$ \label{algline::initialization}
\STATE \textbf{Initialize} $\beta \gets \beta_{0}$
\STATE $t_{noise, i} \gets 1$ \textbf{for} i = 1, 2, ..., l
\FOR{$t = 1, 2, ..., T$}
\STATE $x \gets x - \alpha_{x} \cdot \nabla_{x} L^{*}$
\FOR{i = 1, 2, ..., l }
\IF{$t$ $-$ $t_{noise, i} \geq T_{p}$ and $\|\nabla_{\eta^{(i)}} L^{*})\| \leq \delta$} \label{algline::check_grad_norm}
\STATE $\eta^{(i)} \gets \eta^{(i)} - \alpha_{\eta} \cdot (\nabla_{\eta^{(i)}} L^{*} + r \cdot \xi_{t})$ \label{algline::add_perturbation}
\STATE $t_{noise, i} \gets t$
\ELSE
\STATE $\eta^{(i)} \gets \eta^{(i)} - \alpha_{\eta} \cdot \nabla_{\eta^{(i)}} L^{*}$
\ENDIF
\ENDFOR
\IF{$\|\nabla_{\beta} L^{*}\|$ $\leq$ $\delta_{\beta}$} \label{algline::beta_opt_begin}
\STATE $\beta \gets \beta - \gamma$
\ELSE
\STATE $\beta \gets \beta + \alpha_{\beta} \cdot \nabla_{\beta}L^{*}$
\ENDIF \label{algline::beta_opt_end}
\ENDFOR
\STATE \textbf{Return} $x$
\end{algorithmic}
\end{algorithm}

We evaluate the proposed method in constructing adversarial examples~\cite{goodfellow2014explaining} for neural image classifiers.
Consider a neural image classifier ${C}: {\mathbb{R}^{n} \rightarrow \mathbb{R}^{m}}$, a classification loss ${J}: {\mathbb{R}^{m} \rightarrow \mathbb{R}}$, and a perturbation size ${\epsilon}$. 
For an input image ${x} {\in \mathbb{R}^{n}}$, we derive the perturbation ${\delta}$ that leads the perturbed image 
to maximize the loss ${J}$ within the ${\epsilon}$-neighborhood around ${x}$, i.e., $\argmax_{\|\delta\| \leq \epsilon} {J}({C}({x} + {\delta}))$. 
Two types of adversarial attack are considered: \textit{untargeted} attacks try to misguide the classifier to predict any of the incorrect classes, while \textit{targeted} attacks aim for a particular class.
${J}$ is cross-entropy loss with respect to true label for untargeted attacks, and the negation of cross-entropy loss with respect to target label for targeted attacks. 
We compare the proposed method to Fast Gradient Sign Method (FGSM)~\cite{goodfellow2014explaining} and Projected Gradient Descent (PGD)~\cite{madry2017towards}, two gradient-based attack methods that leverage the signs of input gradients instead of strictly following the steepest descent directions.

\begin{figure*}[htp]
    \centering
    \subfigure[Attack success rates]{           
\includegraphics[width=0.3\linewidth,valign=b]{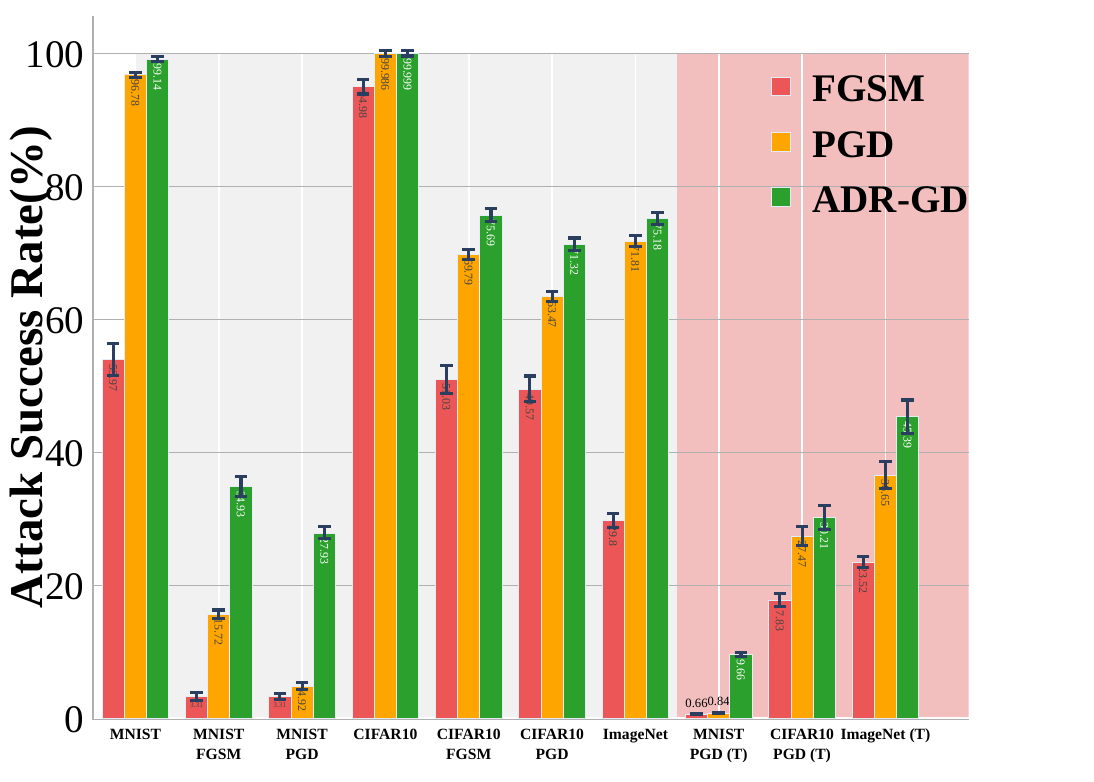} 
        }
    \subfigure[Optimized objective values]{           
        \includegraphics[width=0.3\linewidth,valign=b]{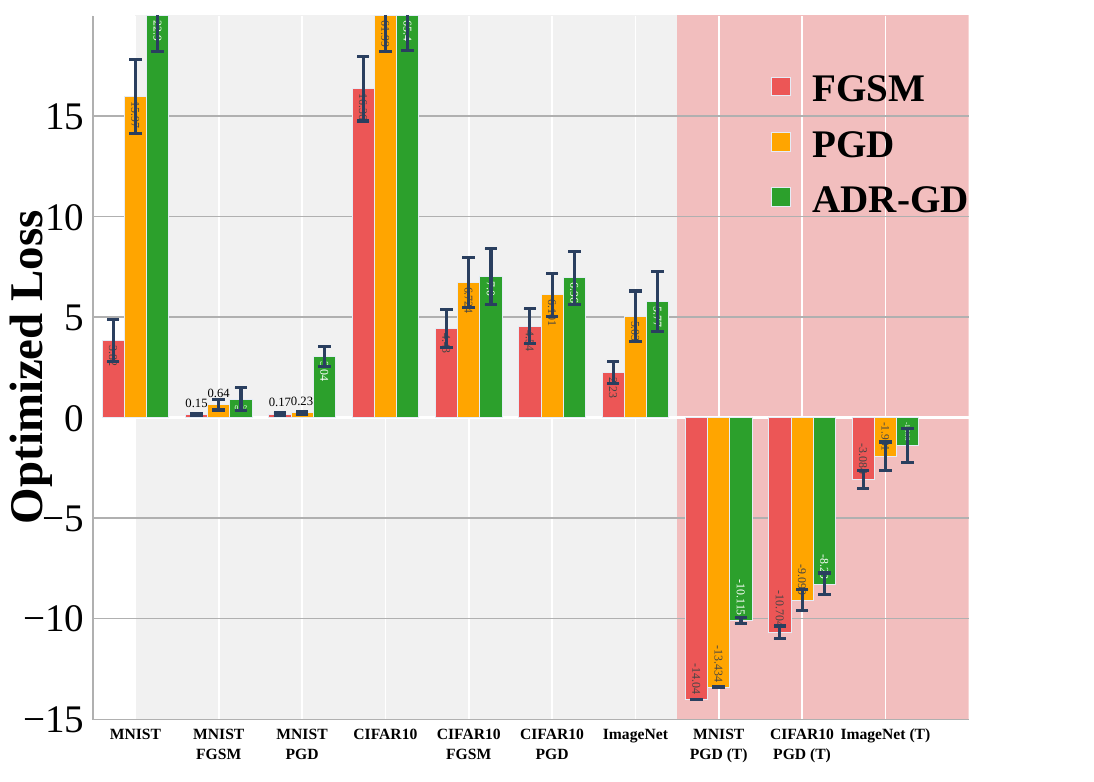} 
        }     
    \subfigure[Episode returns]{
        \includegraphics[width=0.3\linewidth,valign=b]{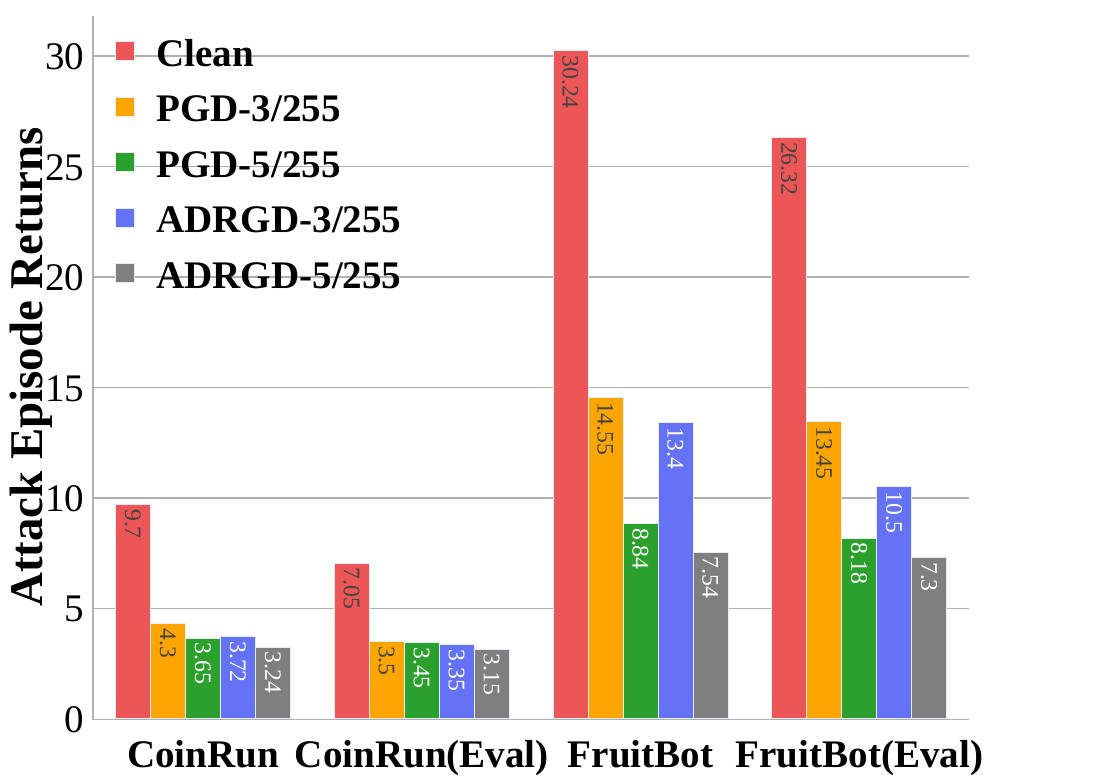} 
    } 
    \caption {\small \textbf{(a)-(b)} Adversarial optimization experiments in Section \ref{section::adversarial_attacking} and \ref{section::attack_drl}. 
    Grey and pink shading indicate the untargeted and targeted experiments respectively. \textbf{(c)} Attacks are constructed to worsen the performance of well-trained neural policies in FruitBot and CoinRun. } 
    \label{fig::adversarial_exp}
\end{figure*}

\begin{figure*}[h!]
    \centering
        \begin{minipage}[b]{0.5\textwidth}
            \centering
            \subfigure[Runtime benchmarks]{
                \adjustbox{valign=b, width=0.9\linewidth}{
                    \begin{tabular}{ccccc}
                    \toprule
                    &\multicolumn{2}{c}{Per-iteration runtime (sec)}   &  \multicolumn{2}{c}{Variable counts}              \\
                    \cmidrule(r){2-3}  
                    \cmidrule(r){4-5}
                    Dataset     & PGD     &  ADR-GD & PGD & ADR-GD   \\
                    \midrule
                    MNIST & $6.22e-4$  & $1.08e-2$ & $784$ & $9180$   \\
                    CIFAR10     & $1.07e-3$ & $1.36e-2$ & $3,072$ & $23,552$     \\
                    ImageNet     & $6.58e-3$       & $4.17e-2$ & $150,528$ & $351,232$ \\
                    \bottomrule
                  \end{tabular}
                }
            } \\ 
            \subfigure[Model architecture]{
                \adjustbox{valign=b, width=0.75\linewidth}{
                    \begin{tabular}{cc}
                    \toprule
                    Model A & {[}{[}10, 64{]}, {[}64, 64{]}, {[}64, 1{]}{]}                     \\ 
                    Model B & {[}{[}10, 500{]}, {[}500, 500{]}, {[}500, 500{]}, {[}500, 1{]}{]}                \\ 
                    Model C & {[}{[}128, 500{]}, {[}500, 500{]}, {[}500, 500{]} {[}500, 1{]}{]} \\ 
                    \bottomrule
                    \end{tabular}
                }
            }
        \end{minipage} \quad
    \subfigure[Optimize objective values]{
        \adjustbox{valign=b, width=0.46\linewidth}{
            \begin{tabular}{cccc}
            \toprule
                         & Model A & Model B & Model C  \\ 
            \midrule
            GD           & 40.58   & 3873.76 & 35807.10 \\ 
            Adam & 41.17   & 3914.07 & 36531.20 \\ 
            Adagrad & 41.72   & 3884.90 & 35513.52 \\ 
            Perturbed GD & 40.62   & 3903.20 & 35964.02 \\ 
            ADR-GD (ours)   & \textbf{43.95}   & \textbf{4047.99} & \textbf{37672.04} \\ 
            M1  & 43.33   & 3914.25 & 36609.58 \\ 
            M2 & 43.74   & 3940.10 & 36822.27 \\ 
            M3  & 43.90   & 3995.62 & 36064.52 \\ 
            M4   & 43.13   & 3882.18 & 36036.73 \\
            \bottomrule
            \end{tabular}
        }} 
        \captionof{table}{\small \textbf{(a)} 
        Average \textit{per-iteration} runtime benchmarks for experiments in Section \ref{section::adversarial_attacking}.
        \textbf{(b)-(c)} Model architectures and results for ablation experiments in Section \ref{section::ablation}. For instance, Model A is a 3-layer ReLU network with the input dimension $d_{n} = 10$, the latent dimensions $d_{1} = d_{2} = 64$, and the output dimension $d_{m} = 1$.}
        \label{table::runtime_and_ablation}
\end{figure*}

\setcounter{subfigure}{0}
\begin{figure*}[htb]
    \centering
    \centering
    \subfigure[\small Ant-v3]{
 \includegraphics[width=0.3\linewidth,valign=b]{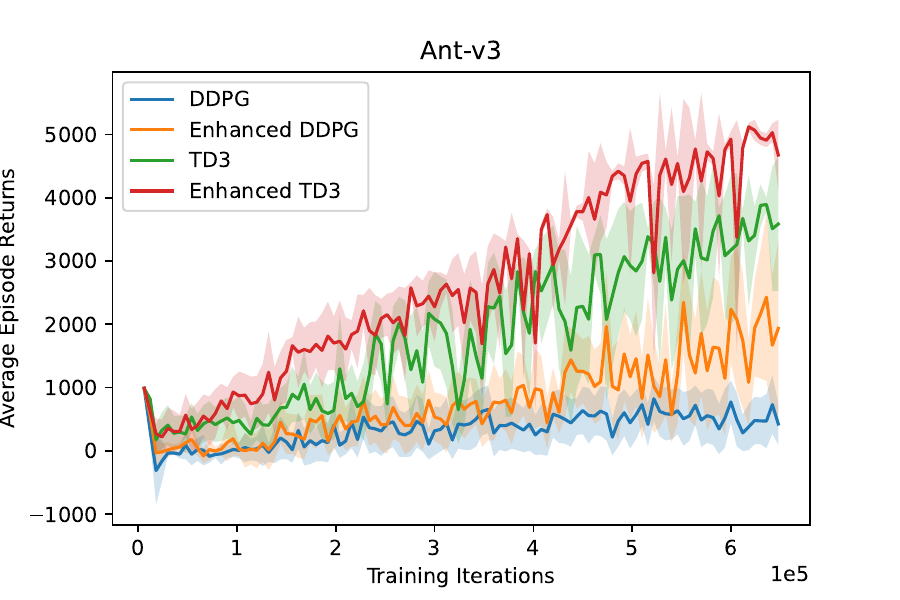} 
    } 
    \subfigure[\small HalfCheetah-v3]{
\includegraphics[width=0.3\linewidth,valign=b]{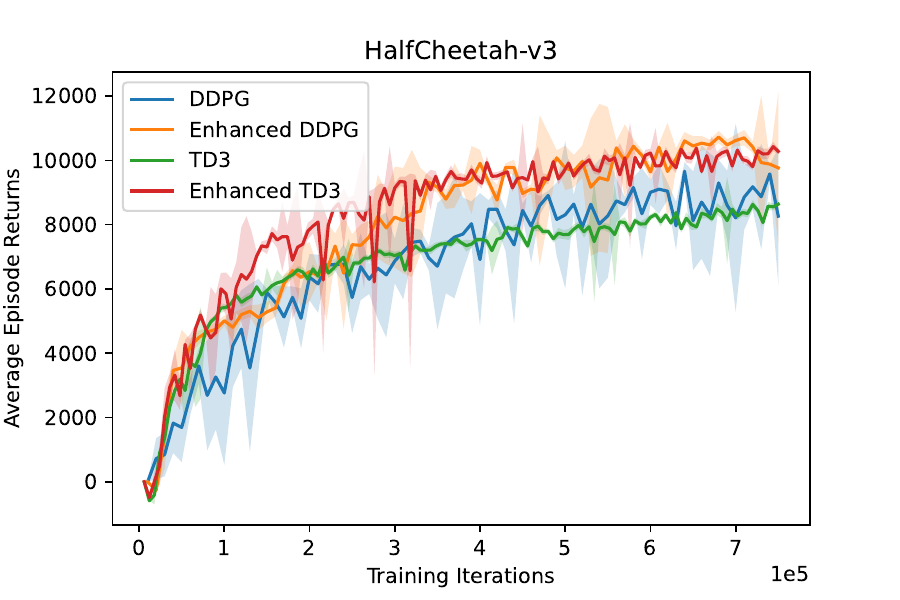} 
    } 
    \subfigure[\small Walker2d-v2]{
\includegraphics[width=0.3\linewidth,valign=b]{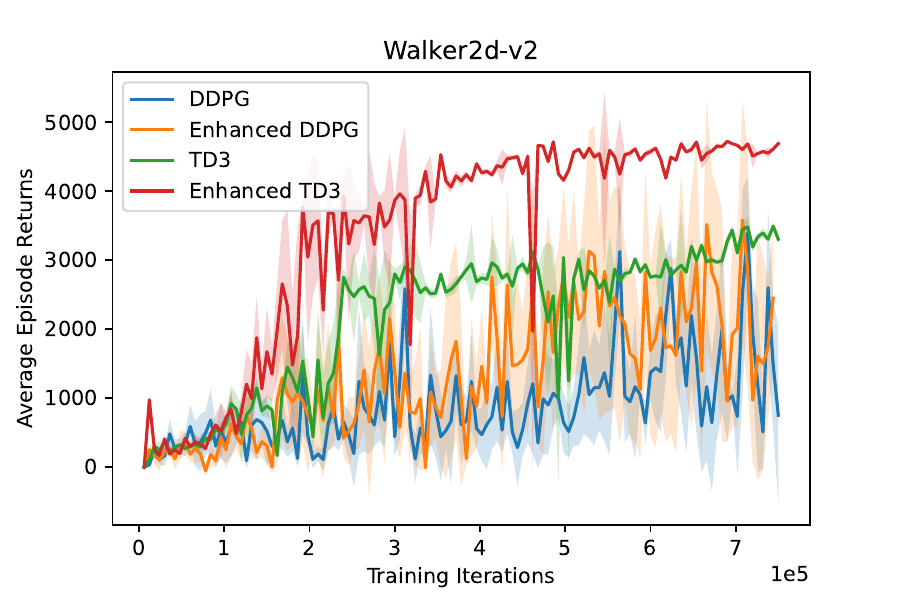} 
    }
    \caption{\small Enhanced DRL training in three mujoco environments}
    \label{fig::drl_exp}
\end{figure*}

We evaluate the proposed algorithm on MNIST~\cite{deng2012mnist}, CIFAR10~\cite{krizhevsky2009learning}, and ImageNet~\cite{deng2009imagenet} datasets.
For MNIST and CIFAR10 datasets, in addition to the standard classification model trained over clean images, we consider the robust models from adversarial training~\cite{shafahi2019adversarial} that use FGSM or PGD methods to generate adversarial examples as supplementary training data.
We employ small-sized Convolutional neural networks for MNIST, while VGG19~\cite{sengupta2019going} for CIFAR10.
For ImageNet, we use the pre-trained Wide ResNet-50-2 model~\cite{zagoruyko2016wide} downloaded from \cite{torchvision2016}.
When dealing with large networks, we can apply the proposed method to partial networks selectively. 
As an example, only the pattern variables over the last layer of Wide ResNet-50-2 model are optimized, while we ignore the neurons from earlier layers just as in vanilla GD.

Figure \ref{fig::adversarial_exp} (a)-(b) summarize the experiment results. 
The perturbation size ${\epsilon}$ is set to $0.2$, $8/255$, and $2/255$ for the evaluations in MNIST, CIFAR10 and ImageNet respectively. 
When evaluating PGD, we allow it to run for a sufficient number of iterations to ensure convergence.  
As shown, the proposed method achieves better attack success-rates and higher optimized loss than PGD and FGSM in all cases.
For all three datasets, PGD delivers comparable performance with the proposed method over the naturally trained models. 
Meanwhile, ADR-GD significantly outperforms PGD on robust models.
For instance, the adversarial training strategy greatly improves the model robustness against PGD attacks on MNIST, reducing the attack success rate of PGD from $96.78\%$ to $4.92\%$.
In comparison, the proposed method achieves $27.93\%$ attack success rate on the robust model.
Moreover, we demonstrate the effectiveness of the proposed method for constructing targeted attacks in  Figure \ref{fig::adversarial_exp}.

Table \ref{table::runtime_and_ablation} (a) provides the average runtime benchmarks for experiments on different datasets.
It shows that the average per-iteration runtime from ADR-GD is slower by nearly an order of magnitude than that from PGD.
This is because the proposed method optimizes more variables, i.e., activation variables $\eta$, than PGD.
Despite the computation overheads from our implementations, we show that computation cost for the proposed algorithm does not grow exponentially with the increase in the network size.

\subsection{Deep Reinforcement Learning}
\label{section::enhanced_drl}

We enhance the training of Deep Reinforcement Learning (DRL) by formulating input optimization problems to compute action refinements during exploitation.
We consider the DRL algorithms that are based on actor-critic structure such as DDPG~\cite{lillicrap2015continuous} and TD3~\cite{fujimoto2018addressing}.
As a motivation, one well-known challenge in these algorithms are to ensure the optimal coupling between actor and critic, i.e.,  actor model properly captures the actions maximizing the approximated Q-values from critic model.

Next, we discuss the formulation of input optimization problem. 
Consider the actor model ${\pi:\mathbb{R}^{n} \rightarrow \mathbb{R}^{m}}$ that maps the observation states to the actions, and the critic model ${V}: {\mathbb{R}^{n + m} \rightarrow \mathbb{R}}$ that approximates the optimal Q-values from state-action pairs. 
We do not modify the training objective of model parameters.
Instead, we compute action refinements by optimizing around the action candidates from actor to maximize the output of critic. 
For an arbitrary state ${s} \in {\mathbb{R}^{n}}$ and a probability threshold ${\epsilon \in [0, 1]}$, we obtain the exploitation action ${a}$ as the following:
\begin{align}
    a = \begin{cases}
        \pi(s)  \hspace{3.2cm} \text{with probability $1 - \epsilon$,}\\
        \pi(s) + \argmax_{(\partial) \in \mathbb{R}^{m}} V(s, \pi(s) + \partial) \hspace{0.2cm} \text{otherwise.}
    \end{cases}
\end{align}
Figure \ref{fig::drl_exp} demonstrates the experiments on several mujoco environments~\cite{todorov2012mujoco}. 
For each testing environment, we apply the proposed enhancement procedures to both DDPG and TD3 algorithms, and use ${\epsilon = 0.25}$ for the enhanced algorithms.
We observe robust improvements from the enhanced methods.
Many prior works reported that a DDPG agent cannot be trained to exceed more than $1,000$ average episode rewards in Ant environment~\cite{fujimoto2018addressing, dankwa2019twin}.
In Figure \ref{fig::drl_exp}, we show that the enhanced DDPG approaches nearly $2,000$ average episode rewards within limited training iterations. 

\subsection{Adversarial Optimization against Neural Policies}
\label{section::attack_drl}

Adversarial attacks to neural policies can be constructed via input optimization.
We consider the environments that have imagery observation space, and discrete action space as in \cite{oikarinen2021robust}. We perform optimization in the observation space, i.e. the policy model's input space, so that non-optimal actions will be produced.
For an arbitrary state ${s} {\in \mathbb{R}^{n}}$, a policy model ${\pi}: {\mathbb{R}^{n} \rightarrow \mathbb{R}^{m}}$ and a classification loss $J: {\mathbb{R}^{m} \rightarrow \mathbb{R}}$ that is minimized at the (sub-)optimal action ${\pi(s)}$, we derive the perturbed state within the ${\epsilon}$-neighborhood of ${x}$ to maximize ${J}$, i.e. ${\argmax_{\|\partial\| \leq \epsilon} J(\pi(s + \partial))}$.
 
Figure \ref{fig::adversarial_exp}(c) demonstrates the evaluations in two Procgen environments~\cite{cobbe2019procgen}, FruitBot and CoinRun. 
We experiment with two attack sizes, i.e., ${\epsilon = 3/255}$ and ${\epsilon = 5/255}$.
Both PGD and the proposed method can construct impactful attacks that sharply undermine the performance of well-trained policies. 
Nevertheless, ADR-GD robustly leads to more significant reductions in average episode rewards than PGD.
Note that in CoinRun environment, the attacks from the proposed method constrained with ${\epsilon = 3/255}$  are as strong as the PGD attacks constrained with ${\epsilon = 5/255}$.

\subsection{Ablation Study}
\label{section::ablation}
We perform ablation experiment to illustrate the importance of each components in Algorithm \ref{alg::proposed}.We test with randomized neural networks that produce scalar outcomes. Model parameters are uniformly sampled from range $(-1, 1)$. Denote one randomized network with $F: {\mathbb{R}^{n} \rightarrow \mathbb{R}}$. The problem objective is simply to maximize the network's scalar output, i.e., ${\max_{x \in [-1, 1]^{n}} F(x)}$. 

Table \ref{table::runtime_and_ablation}(b) provides the model architectures of testing models. To understand how each component contributes to Algorithm \ref{alg::proposed}, we experiment with different settings as: \textbf{M1} - detaching the gradient of $J(\bar{h}^{(l+1)}(x, \eta))$ on $\eta$, \textbf{M2} - detaching the gradient of $\beta \sum_{i=1}^{n-1} L_{i}$ on $x$,  \textbf{M3} - removing normalization, and   \textbf{M4} - removing perturbation.
Table \ref{table::runtime_and_ablation}(c) shows the average objective value ${F(x^{*})}$ where ${x^*}$ is the optimized input variable. 
The proposed method with all the components enabled always achieves the highest ${F(x^{*})}$. 
Decays in the optimized value are observed when we detach the gradients of partial objectives (e.g. \textbf{M1} and \textbf{M2}). 
We note that \textbf{M3} delivers comparable performance with the proposed method for small networks. It is reasonable as for small networks, the magnitude order of hidden neurons may not differ much at different layers, and thus normalization is not necessary. 
The performance of Perturbed GD is limited as expected, due to the piecewise gradient landscape in deep ReLU networks. 
Nevertheless, it is demonstrated the improvement in the proposed algorithm led by introducing randomized perturbations to the activation pattern variables.

\section{Conclusion}

We present a novel optimization procedure for input optimization of ReLU Networks that explicitly takes into account the impact of changes in activation patterns on the output. 
We introduced new regularization terms for local descent in the input space that encourages the input change to be aligned with the descent directions in the activation space. The overall procedures can thus achieve better local descent properties than GD and also various forms of randomly perturbed gradient methods. We observed that the proposed methods improve the state-of-the-art results from prior gradient-based optimization methods in various application problems. 

Future work in this direction can involve further theoretical analysis for the convergence complexity and improving scalability to larger networks.  
In addition, we believe the proposed method offers useful insights into optimizing non-ReLU networks whose value landscape is piecewise-continuous, rather than piecewise-linear as in ReLU networks.
Therefore, these networks also suffer from the same issues discussed in Section \ref{section::gd_problems}, when following the standard gradient descent directions over their non-convex landscape. 
\section*{Acknowledgements}

We thank the anonymous reviewers for their valuable suggestions.
This material is based on work supported by NSF Career CCF 2047034, NSF CCF 2112665 TILOS AI Institute, NSF CCF DASS 2217723, and Amazon Research Award.

\section*{Impact Statement}
This paper presents work aiming at solving input optimization problems for ReLU networks.
The negative impact of it is the potential use that perform adversarial optimizations to intentionally attack or disturb well-trained ReLU network models for vicious purposes.
Nevertheless, the development of any adversarial algorithms is to achieve a truly robust neural models that can defend any attacks.  
The optimization criteria of the proposed method takes into consideration the combinatorial nature underneath ReLU networks, which we believe can be related to understanding the expressivity and generalizability of ReLU networks in the future.   

\bibliography{citation}
\bibliographystyle{icml2024}

\appendix
\onecolumn

\section{Approximating Piecewise Constant Function with Sigmoid }
\label{section::sigmoid}

In this section, we discuss the impact of the modified sigmoid used in the sigmoid-surrogate network (\ref{eq::sigmoid_network}). 
Recall that we employ a following variant of sigmoid function with an offset of $1/2$: 
$$s_{\alpha}(x) = \frac{1}{1 + \exp(-\alpha(x - \frac{1}{2}))},$$
where $\alpha$ is the hyperparameter that determines how fast the function value grows from $0$ to $1$.
At high level, we introduce sigmoid functions to relax the binary activation variables, i.e. $\nu$ in (\ref{nndef2}),  into continuous representations, i.e. $\eta$ in (\ref{eq::sigmoid_network}).
Next, we illustrate how the approximation errors from the proposed relaxation does not worsen the optimization on input variables, as long as the value of hyperparameter $\alpha$ is well selected. 

First, we briefly revisit the ReLU operator, and rewrite its procedures as follows:
\begin{align*}
    \sigma(x) &= \max(0, 1)\\ 
    &= \mathbbm{1}(x>0) \cdot x
\end{align*}
which shows that the ReLU operator essentially derives a binary coefficient, i.e. $\mathbbm{1}(x>0)$, whose value depends on if $x$ is positive or not.

\begin{figure}[htp]
  \centering
  \subfigure{
      \includegraphics[height=3.2cm, valign=m]{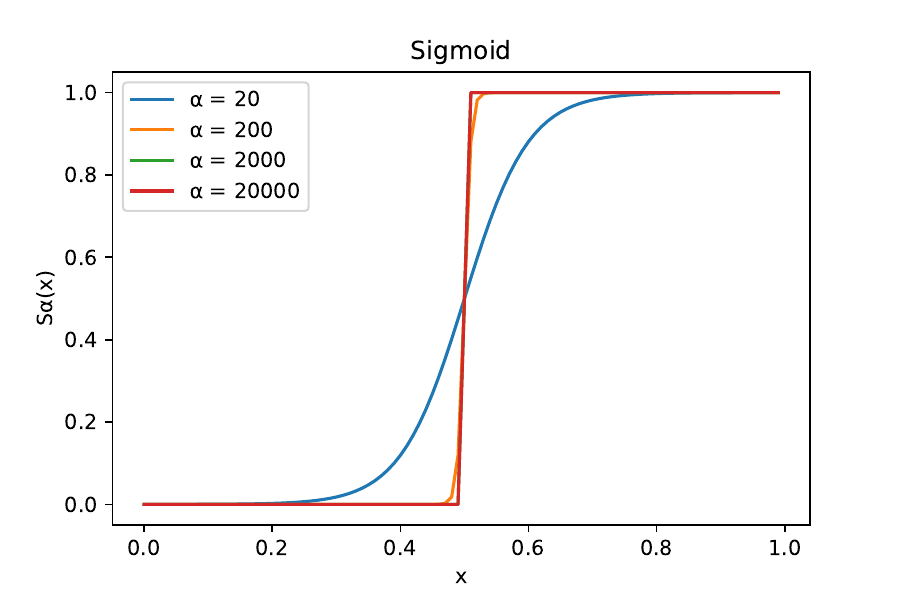}
  }
  \subfigure{
        \adjustbox{valign=m, width=0.4\textwidth}{
            \begin{tabular}{c|c|c|c}
                         & Model A & Model C  \\ \hline
            GD           & 40.58   &  35807.10 \\ \hline
            Ours,  $\alpha = 20$  &  25.84 & 19942.94 \\ \hline
            Ours,  $\alpha = 200$ &  43.75 & 36219.05 \\ \hline
            Ours,  $\alpha = 2000$  & 43.95   & \textbf{37672.04} \\ \hline
            Ours,  $\alpha = 20000$   & \textbf{44.12}  & 37054.31
            \end{tabular}
    }}
    \caption{\small\textbf{Left}: Visualization of sigmoid functions under different values of $\alpha$.\textbf{Right}: The average \textit{optimized} objective values for the models in Table 1.}
    \label{fig::sigmoid_ablation}
\end{figure}

Figure \ref{fig::sigmoid_ablation}(Left) visualizes a 1-D sigmoid function within input range $[0, 1]$, using different values of $\alpha$. 
It shows that as the value of $\alpha$ arises, $s_{\alpha}(x)$ approaches a piecewise constant function bounded between $[0, 1]$, just like the ReLU operator. 
An overlarge value of $\alpha$ may undermine our incentive to relax the ReLU operator continuously, while an oversmall $\alpha$ value introduces high approximation errors with respect to the ReLU operator. 
Figure \ref{fig::sigmoid_ablation}(Right) demonstrates the ablation experiment with varying values of $\alpha$.
We evaluate on randomized neural networks that follow the same architectures with those in Table \ref{table::runtime_and_ablation}(a).
For the experiment results reported in main paper, we employ $\alpha = 2000$.

\section{Proof of Proposition \ref{prop::eta_gradient}}
\label{appendix::proof}

In this section, we show that the gradient of the output of sigmoid-surrogate network in (\ref{eq::sigmoid_network}) over the activation variables $\eta$ is consistent with the steepest direction in Definition~\ref{steepest}.
We achieve this by showing that the gradient of the surrogate function over $\eta$ always forms a positive inner-product with the steepest direction in the discrete form.


\begin{proof}

Let $\bar{h}^{(i+1)}_{k}(x, \eta)$ be an arbitrary component of the sigmoid-surrogate network defined in (\ref{eq::sigmoid_network}). We derive the gradient of $\bar{h}^{(i+1)}_{k}(x, \eta)$ on $\eta^{(i)}_{j}$ as:
\begin{eqnarray}
    \partial \eta^{(i)}_{j} = \frac{\partial \bar{h}^{(i+1)}_{k} (x, \eta)}{\partial s_{\alpha}(\eta^{(i)}_{j})} \cdot \frac{1}{\bar{h}^{(i)}_j(x, \eta)} \cdot s_{\alpha}(\eta^{(i)}_{j}) \cdot (1 - s_{\alpha}(\eta^{(i)}_{j})) \cdot \alpha
\end{eqnarray}
Next, we show that $\partial \eta^{(i)}_{j}$ always forms a positive product with $\partial \nu^{(i)}_{j}$ defined in (\ref{steepest}). 
If $\partial \nu^{(i)}_{j} = 0$, then $\partial \nu^{(i)}_{j} \cdot \partial \eta^{(i)}_{j} \geq 0$ must be trivially true. If $\partial \nu^{(i)}_{j} \neq 0$ and $\nu^{(i)}_{j} = 1$, then:
\begin{align*}
    \partial \nu^{(i)}_{j} &= \displaystyle\hat{\sgn}\bigg(\frac{\partial h^{(l+1)}_k(x,\nu)}{\partial \nu^{(i)}_{j}}\cdot \frac{1}{h^{(i)}_j(x,\nu)}\bigg) \\
\end{align*}
As discussed in Section \ref{section::sigmoid}, $s_{\alpha}(\eta^{(i)}_j)$ and $\bar{h}^{(i)}_j(x, \eta)$ closely approach $\nu^{(i)}_{j}$ and $h^{((i))}_j(x,\nu)$ respectively, as long as $\eta^{(i)}_j$ and $\nu^{(i)}_j$ both satisfy the feasibility constraints in (\ref{feasible}) and (\ref{feasible2}).
In fact, it must hold true that:
\begin{align*}
    \hat{\sgn}(\partial \eta^{(i)}_{j}) &= \partial \nu^{(i)}_{j}
\end{align*}
This is because $s_{\alpha}(\eta^{(i)}_{j}) \cdot (1 - s_{\alpha}(\eta^{(i)}_{j}))$ must always yield a non-negative value for all possible values of $\eta^{(i)}_{j}$.
Thus it must be true that $\partial \nu^{(i)}_{j} \cdot \partial \eta^{(i)}_{j} \geq 0$.
The case for $\nu^{(i)}_{j} = 0$ can be proved similarly.
\end{proof}

\section{Experiment Parameters}
\label{appendix::experiment_parameters}

Note that we employ gradient normalization in our implementations for optimizing $x$ and $\eta$, in case of any concerns about the high learning-rates. 

\begin{figure}[h]
    \centering
    \subfigure[Untargeted attacks in Section \ref{section::adversarial_attacking}]{
    \adjustbox{valign=b, height=3.5cm}{
        \begin{tabular}{cccc}
            \toprule
            & MNIST & CIFAR10 & ImageNet   \\ 
            \midrule
            T                             & 250   & 500  & 500      \\ 
            $\beta_{0}$                    & 2.   & 0.5 & 0.5   \\ 
            $\alpha_x$                    & 1.   & 0.1 & 0.1 \\ 
            $\alpha_{\sigma}$  & 0.25  & 0.5 & 0.5      \\ 
            $\alpha_\beta$          & 0.001    & 0.001  & 0.001       \\ 
            r                             & 0.1    & 0.2 & 0.1       \\ 
            $\delta$ & 0.0001  & 0.001 & 0.01  \\ 
            $\delta_{\beta}$     & 0.1  & 0.1 & 0.05    \\ 
            $T_{p}$                & 5    & 5  & 5       \\ 
            \bottomrule
        \end{tabular}
    }} \quad
    \subfigure[Targeted attacks in Section \ref{section::adversarial_attacking}]{
    \adjustbox{valign=b, height=3.5cm}{
        \begin{tabular}{cccc}
            \toprule
            & MNIST & CIFAR10 & ImageNet    \\ 
            \midrule
            T                             & 250   & 500 & 500       \\ 
            $\beta_{0}$                    & 2.   & 1. & 0.5  \\
            $\alpha_x$                    & 0.5   & 0.1 & 0.1 \\
            $\alpha_{\sigma}$  & 0.1  & 0.5  & 0.5    \\
            $\alpha_\beta$          & 0.001    & 0.001  & 0.0001      \\
            r                             & 0.01    & 0.1  & 0.01      \\
            $\delta$ & 0.0001  & 0.001  & 0.01\\ 
            $\delta_{\beta}$     & 0.1  & 0.05 & 0.05     \\ 
            $T_{p}$                & 5    & 5    & 5     \\
            \bottomrule
        \end{tabular}
    }}\\
    \subfigure[Enhanced DRL training in Section \ref{section::enhanced_drl}]{
    \adjustbox{valign=b, height=3.5cm}{
        \begin{tabular}{ccc}
            \toprule
            & DDPG & TD3 \\ 
            \midrule
            T          & 100    & 100     \\ 
            $\beta_{0}$        & 0.5   & 0.5  \\ 
            $\alpha_x$     & 0.2   & 0.2    \\ 
            $\alpha_{\sigma}$  & 0.1    & 0.1      \\ 
            $\alpha_\beta$          & 0.01    & 0.01        \\ 
            r          & 0.05    & 0.05     \\ 
            $\delta$ & 0.001  & 0.001  \\ 
            $\delta_{\beta}$     & 0.1  & 0.1     \\ 
            $T_{perturb}$ & 10      & 10       \\ 
            \bottomrule
        \end{tabular}
    }} \quad
    \subfigure[Attacking DRL policies in Section \ref{section::attack_drl}]{
    \adjustbox{valign=b, height=3.5cm}{
        \begin{tabular}{ccc}
            \toprule
            & FruitBot & CoinRun \\ 
            \midrule
            T          & 100    & 100     \\ 
            $\beta_{0}$        & 0.5   & 0.5  \\ 
            $\alpha_x$     & 0.4   & 0.4    \\ 
            $\alpha_{\sigma}$  & 0.25    & 0.25      \\ 
            $\alpha_\beta$          & 0.001    & 0.001        \\ 
            r          & 0.01    & 0.01     \\ 
            $\delta$ & 0.0001  & 0.0001  \\ 
            $\delta_{\beta}$     & 0.1  & 0.1     \\ 
            $T_{perturb}$ & 10      & 10       \\ 
            \bottomrule
        \end{tabular}
    }}
    \subfigure[Ablation in Section \ref{section::ablation}]{
    \adjustbox{valign=b, height=3.5cm}{
    \begin{tabular}{cccc}
        \toprule
        & Model A & Model B & Model C \\ 
        \midrule
        T          & 3000    & 3000    & 3000    \\ 
        $\beta_0$       & 1.      & 1.      & 1.      \\ 
        $\alpha_x$      & 0.5   & 0.5   & 0.5  \\ 
        $\alpha_{\sigma}$  & 1.     & 1.     & 1.25     \\ 
        $\alpha_{\beta}$       & 0.01      & 0.005      & 0.0001      \\ 
        r          & 0.1     & 0.2     & 0.35     \\ 
        $\delta$     & 0.0001  & 0.001  & 0.01  \\ 
        $\delta_{\beta}$     & 0.001  & 0.01  & 0.01  \\ 
        $T_{p}$ & 25      & 25      & 25      \\ 
        \bottomrule
    \end{tabular}
    }}
    \captionof{table}{Training parameters}
\end{figure}

\section{Visualizations of the Perturbed Images}
\label{appendix::visuals}

\begin{figure}
    \centering
    \includegraphics{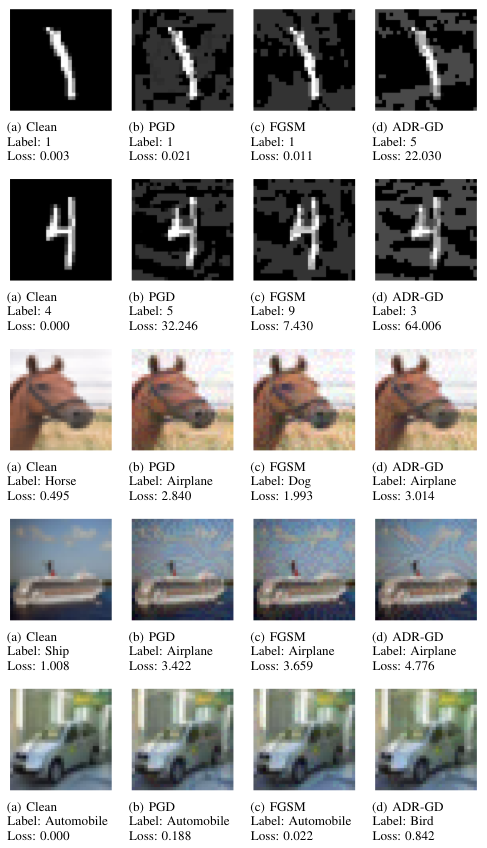}
\end{figure}

\begin{figure}
    \centering
    \includegraphics{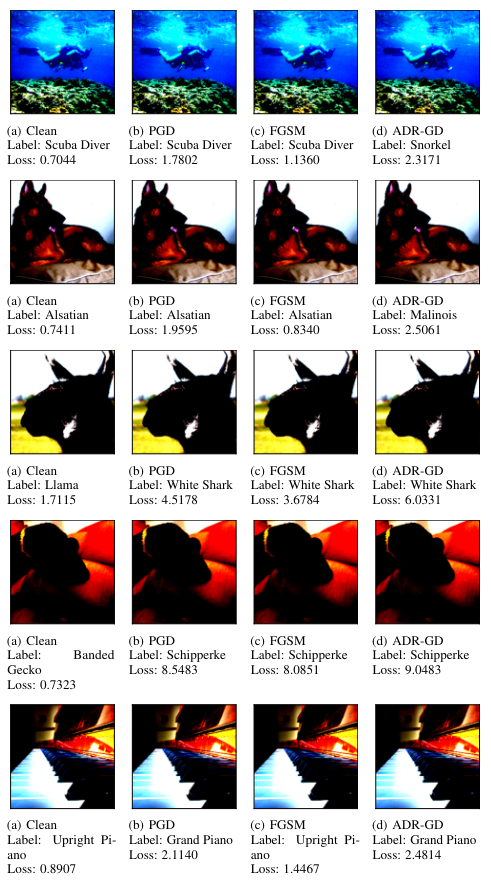}
\end{figure}

\begin{figure}[h]
  \centering
  \subfigure{\includegraphics[width=0.12\textwidth]{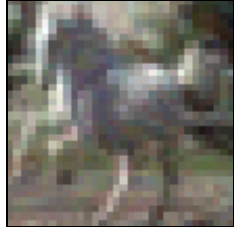}}
  \subfigure{\includegraphics[width=0.12\textwidth]{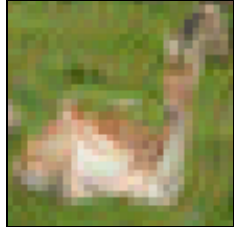}}
  \subfigure{\includegraphics[width=0.12\textwidth]{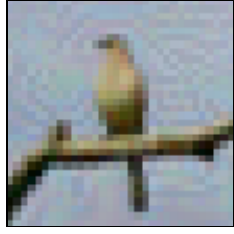}} 
  \subfigure{\includegraphics[width=0.12\textwidth]{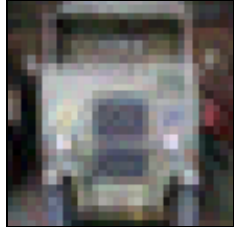}}
  \subfigure{\includegraphics[width=0.12\textwidth]{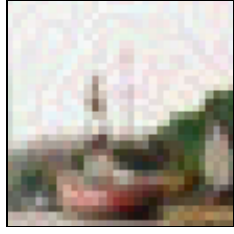}}
  \subfigure{\includegraphics[width=0.12\textwidth]{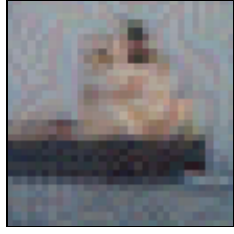}} 
  \subfigure{\includegraphics[width=0.12\textwidth]{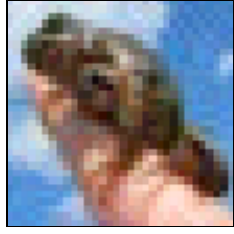}} \\
  \subfigure{\includegraphics[width=0.12\textwidth]{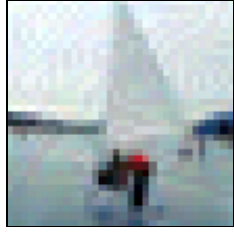}}
  \subfigure{\includegraphics[width=0.12\textwidth]{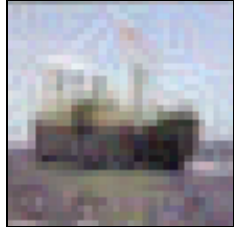}}
  \subfigure{\includegraphics[width=0.12\textwidth]{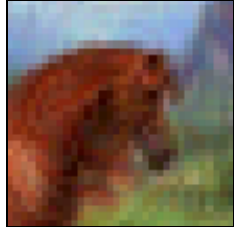}} 
  \subfigure{\includegraphics[width=0.12\textwidth]{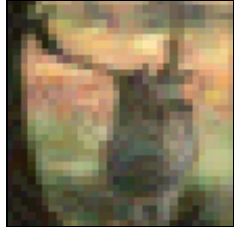}} 
  \subfigure{\includegraphics[width=0.12\textwidth]{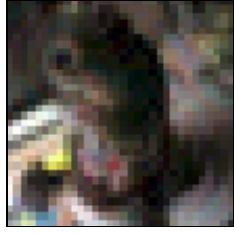}}
  \subfigure{\includegraphics[width=0.12\textwidth]{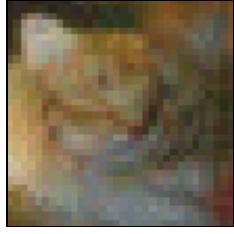}} 
  \subfigure{\includegraphics[width=0.12\textwidth]{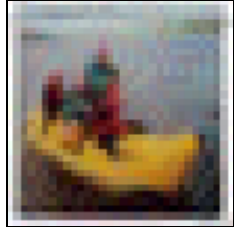}} \\ 
  \subfigure{\includegraphics[width=0.12\textwidth]{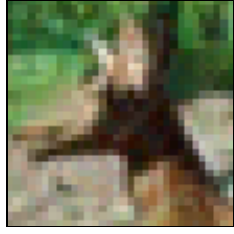}}
  \subfigure{\includegraphics[width=0.12\textwidth]{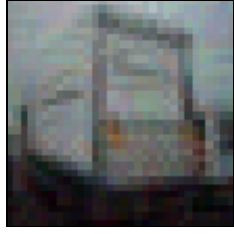}}
  \subfigure{\includegraphics[width=0.12\textwidth]{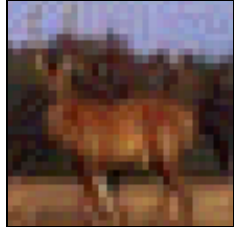}} 
  \subfigure{\includegraphics[width=0.12\textwidth]{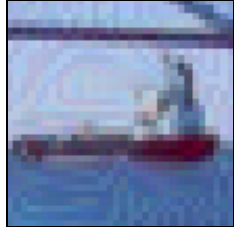}} 
  \subfigure{\includegraphics[width=0.12\textwidth]{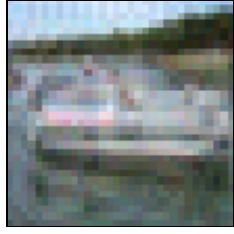}}
  \subfigure{\includegraphics[width=0.12\textwidth]{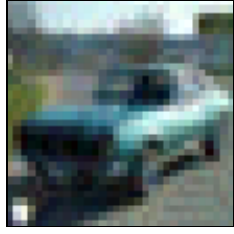}} 
  \subfigure{\includegraphics[width=0.12\textwidth]{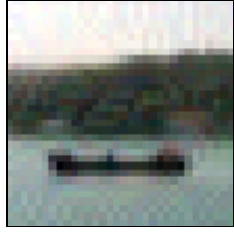}} \\ 
  \subfigure{\includegraphics[width=0.12\textwidth]{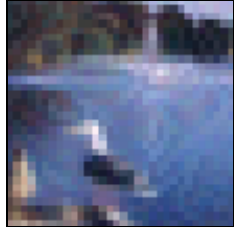}}
  \subfigure{\includegraphics[width=0.12\textwidth]{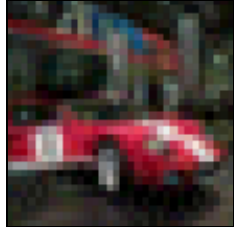}}
  \subfigure{\includegraphics[width=0.12\textwidth]{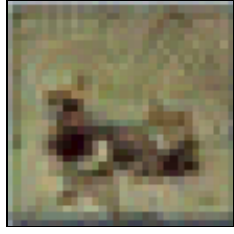}} 
  \subfigure{\includegraphics[width=0.12\textwidth]{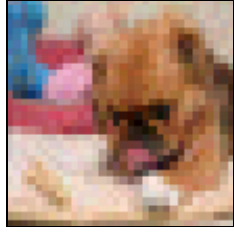}} 
  \subfigure{\includegraphics[width=0.12\textwidth]{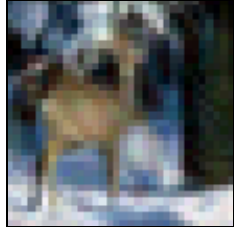}}
  \subfigure{\includegraphics[width=0.12\textwidth]{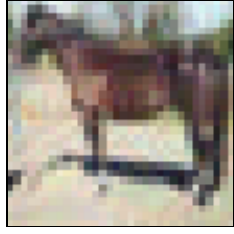}} 
  \subfigure{\includegraphics[width=0.12\textwidth]{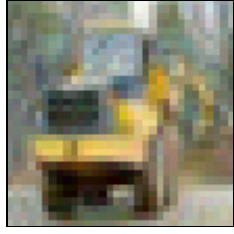}} \\ 
  \subfigure{\includegraphics[width=0.12\textwidth]{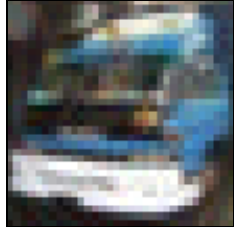}}
  \subfigure{\includegraphics[width=0.12\textwidth]{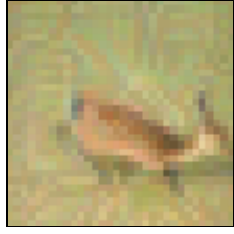}}
  \subfigure{\includegraphics[width=0.12\textwidth]{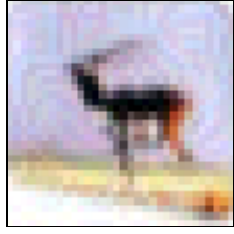}} 
  \subfigure{\includegraphics[width=0.12\textwidth]{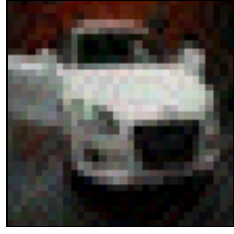}} 
  \subfigure{\includegraphics[width=0.12\textwidth]{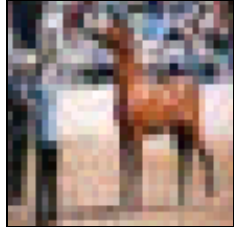}}
  \subfigure{\includegraphics[width=0.12\textwidth]{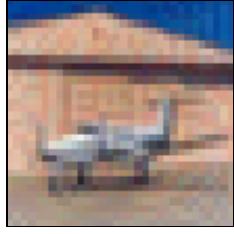}} 
  \subfigure{\includegraphics[width=0.12\textwidth]{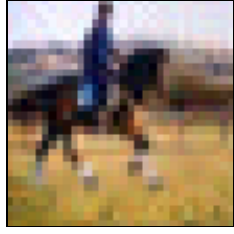}} \\ 
  \subfigure{\includegraphics[width=0.12\textwidth]{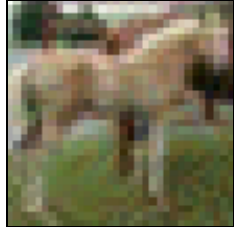}}
  \subfigure{\includegraphics[width=0.12\textwidth]{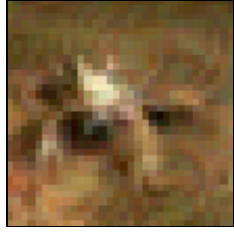}}
  \subfigure{\includegraphics[width=0.12\textwidth]{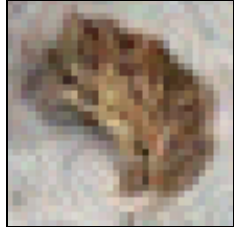}} 
  \subfigure{\includegraphics[width=0.12\textwidth]{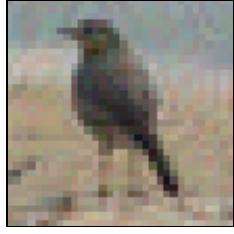}} 
  \subfigure{\includegraphics[width=0.12\textwidth]{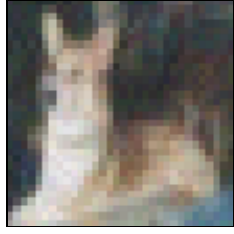}}
  \subfigure{\includegraphics[width=0.12\textwidth]{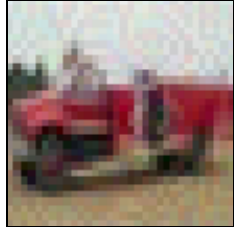}} 
  \subfigure{\includegraphics[width=0.12\textwidth]{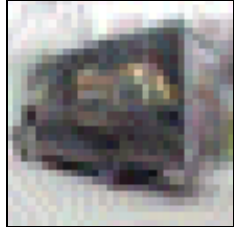}} \\ 
  \subfigure{\includegraphics[width=0.12\textwidth]{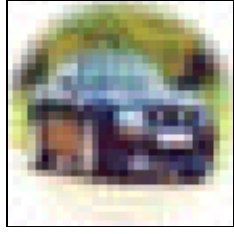}}
  \subfigure{\includegraphics[width=0.12\textwidth]{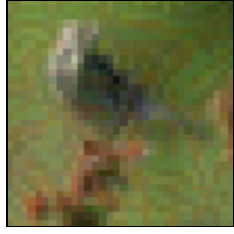}}
  \subfigure{\includegraphics[width=0.12\textwidth]{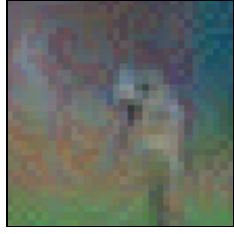}} 
  \subfigure{\includegraphics[width=0.12\textwidth]{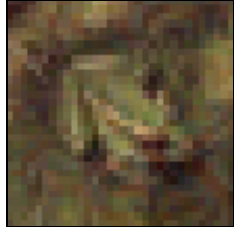}} 
  \subfigure{\includegraphics[width=0.12\textwidth]{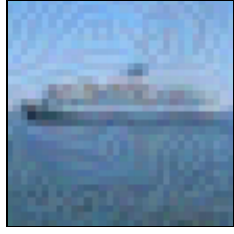}}
  \subfigure{\includegraphics[width=0.12\textwidth]{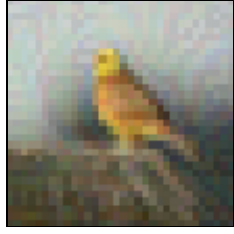}} 
  \subfigure{\includegraphics[width=0.12\textwidth]{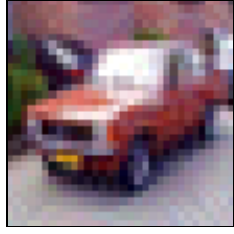}}
  \caption{More adversarial examples constructed by the proposed method for CIFAR10, with $\epsilon = 8/255$. }
\end{figure}

\begin{figure}[h]
  \centering
  \subfigure{\includegraphics[width=0.12\textwidth]{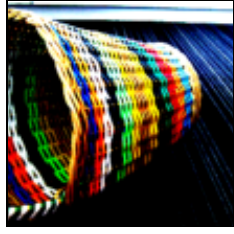}}
  \subfigure{\includegraphics[width=0.12\textwidth]{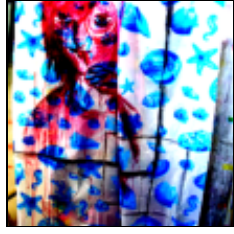}}
  \subfigure{\includegraphics[width=0.12\textwidth]{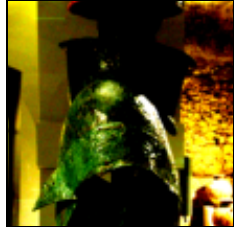}} 
  \subfigure{\includegraphics[width=0.12\textwidth]{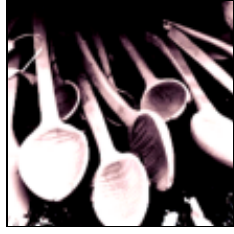}}
  \subfigure{\includegraphics[width=0.12\textwidth]{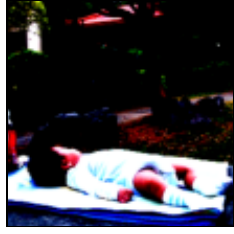}}
  \subfigure{\includegraphics[width=0.12\textwidth]{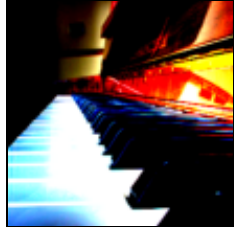}} 
  \subfigure{\includegraphics[width=0.12\textwidth]{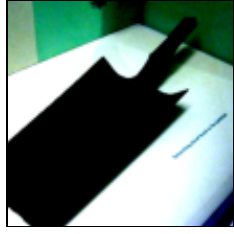}} \\
  \subfigure{\includegraphics[width=0.12\textwidth]{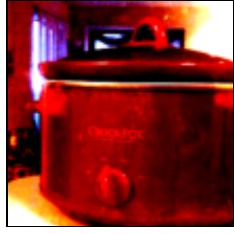}}
  \subfigure{\includegraphics[width=0.12\textwidth]{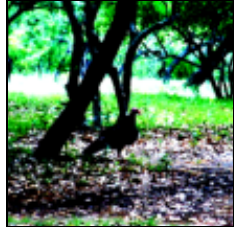}}
  \subfigure{\includegraphics[width=0.12\textwidth]{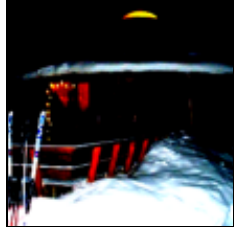}} 
  \subfigure{\includegraphics[width=0.12\textwidth]{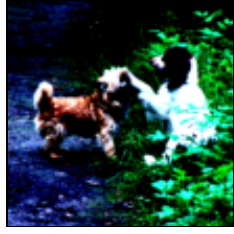}} 
  \subfigure{\includegraphics[width=0.12\textwidth]{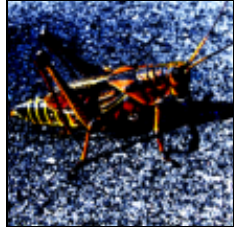}}
  \subfigure{\includegraphics[width=0.12\textwidth]{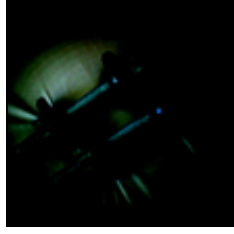}} 
  \subfigure{\includegraphics[width=0.12\textwidth]{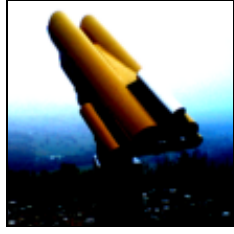}} \\ 
  \subfigure{\includegraphics[width=0.12\textwidth]{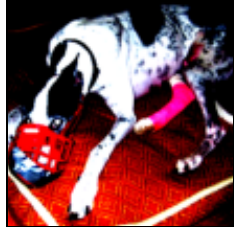}}
  \subfigure{\includegraphics[width=0.12\textwidth]{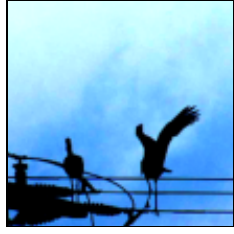}}
  \subfigure{\includegraphics[width=0.12\textwidth]{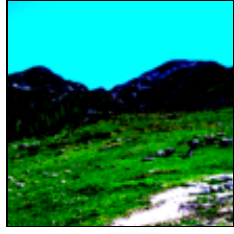}} 
  \subfigure{\includegraphics[width=0.12\textwidth]{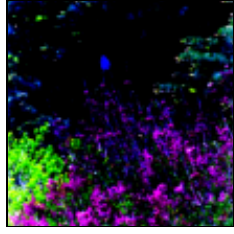}} 
  \subfigure{\includegraphics[width=0.12\textwidth]{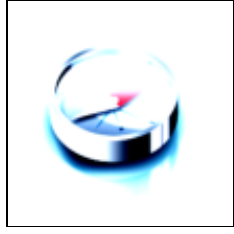}}
  \subfigure{\includegraphics[width=0.12\textwidth]{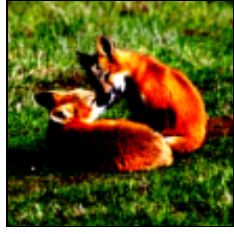}} 
  \subfigure{\includegraphics[width=0.12\textwidth]{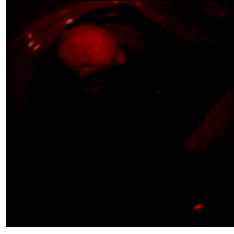}} \\ 
  \subfigure{\includegraphics[width=0.12\textwidth]{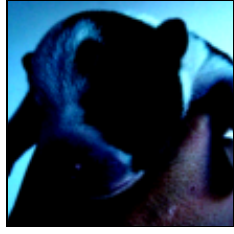}}
  \subfigure{\includegraphics[width=0.12\textwidth]{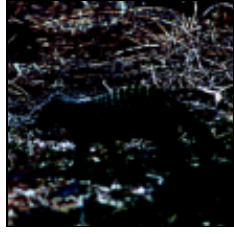}}
  \subfigure{\includegraphics[width=0.12\textwidth]{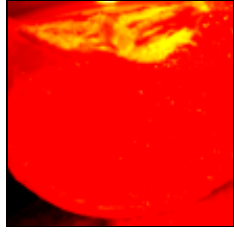}} 
  \subfigure{\includegraphics[width=0.12\textwidth]{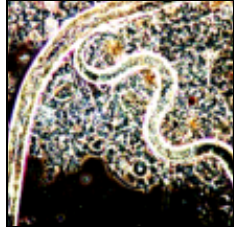}} 
  \subfigure{\includegraphics[width=0.12\textwidth]{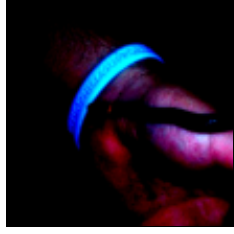}}
  \subfigure{\includegraphics[width=0.12\textwidth]{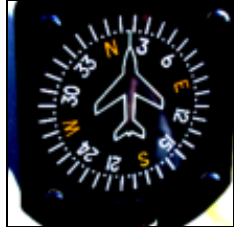}} 
  \subfigure{\includegraphics[width=0.12\textwidth]{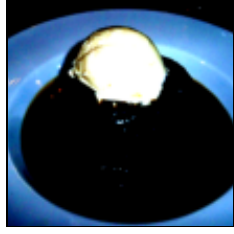}} \\ 
  \subfigure{\includegraphics[width=0.12\textwidth]{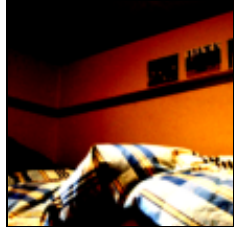}}
  \subfigure{\includegraphics[width=0.12\textwidth]{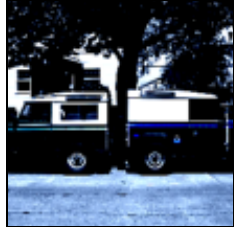}}
  \subfigure{\includegraphics[width=0.12\textwidth]{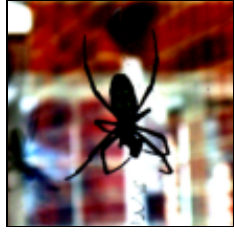}} 
  \subfigure{\includegraphics[width=0.12\textwidth]{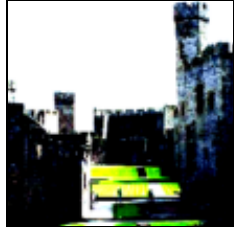}} 
  \subfigure{\includegraphics[width=0.12\textwidth]{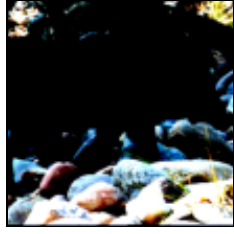}}
  \subfigure{\includegraphics[width=0.12\textwidth]{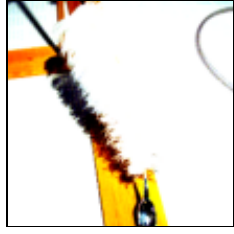}} 
  \subfigure{\includegraphics[width=0.12\textwidth]{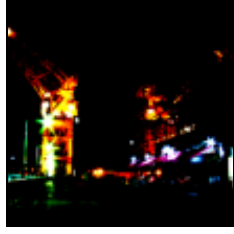}} \\ 
  \subfigure{\includegraphics[width=0.12\textwidth]{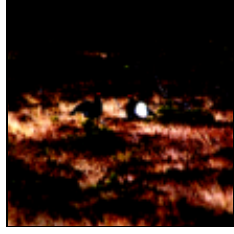}}
  \subfigure{\includegraphics[width=0.12\textwidth]{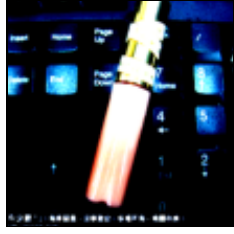}}
  \subfigure{\includegraphics[width=0.12\textwidth]{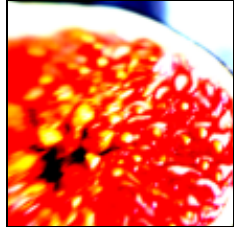}} 
  \subfigure{\includegraphics[width=0.12\textwidth]{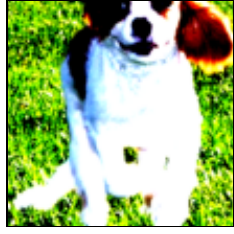}} 
  \subfigure{\includegraphics[width=0.12\textwidth]{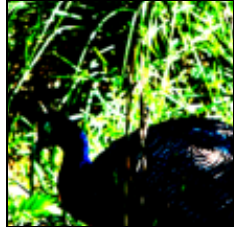}}
  \subfigure{\includegraphics[width=0.12\textwidth]{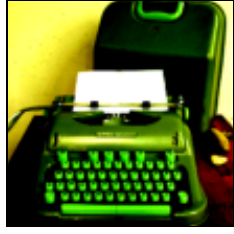}} 
  \subfigure{\includegraphics[width=0.12\textwidth]{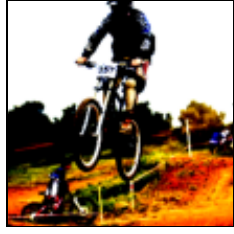}} \\ 
  \subfigure{\includegraphics[width=0.12\textwidth]{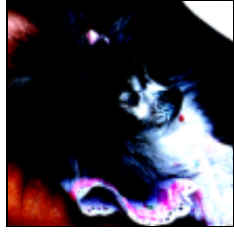}}
  \subfigure{\includegraphics[width=0.12\textwidth]{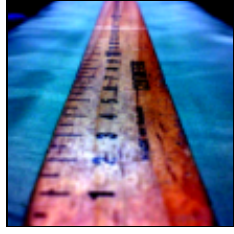}}
  \subfigure{\includegraphics[width=0.12\textwidth]{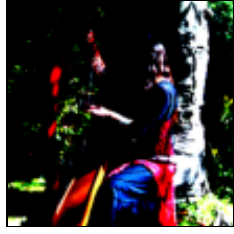}} 
  \subfigure{\includegraphics[width=0.12\textwidth]{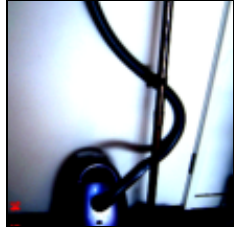}} 
  \subfigure{\includegraphics[width=0.12\textwidth]{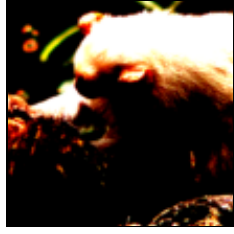}}
  \subfigure{\includegraphics[width=0.12\textwidth]{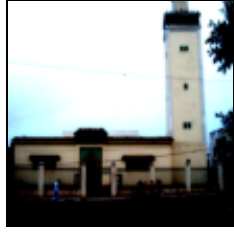}} 
  \subfigure{\includegraphics[width=0.12\textwidth]{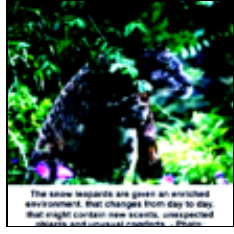}}
  \caption{More adversarial examples constructed by the proposed method for ImageNet, with $\epsilon = 1/255$. }
\end{figure}

\begin{figure}[h]
  \centering
  \subfigure{\includegraphics[width=0.12\textwidth]{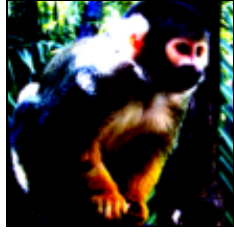}}
  \subfigure{\includegraphics[width=0.12\textwidth]{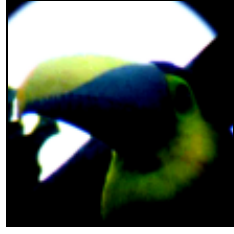}}
  \subfigure{\includegraphics[width=0.12\textwidth]{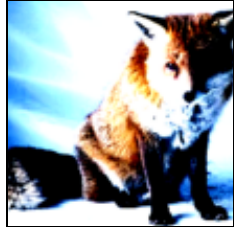}} 
  \subfigure{\includegraphics[width=0.12\textwidth]{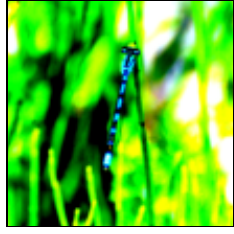}}
  \subfigure{\includegraphics[width=0.12\textwidth]{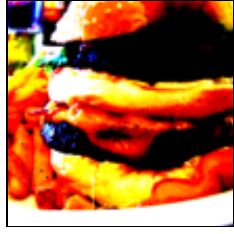}}
  \subfigure{\includegraphics[width=0.12\textwidth]{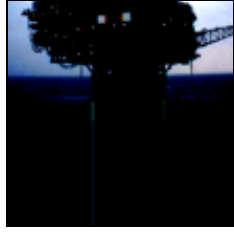}} 
  \subfigure{\includegraphics[width=0.12\textwidth]{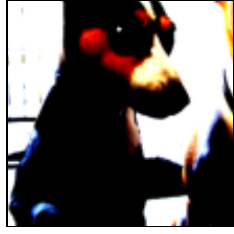}} \\
  \subfigure{\includegraphics[width=0.12\textwidth]{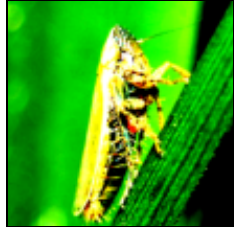}}
  \subfigure{\includegraphics[width=0.12\textwidth]{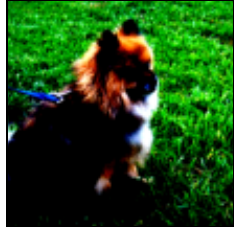}}
  \subfigure{\includegraphics[width=0.12\textwidth]{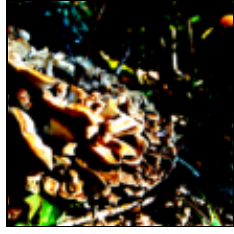}} 
  \subfigure{\includegraphics[width=0.12\textwidth]{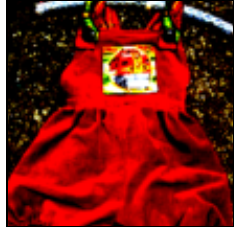}} 
  \subfigure{\includegraphics[width=0.12\textwidth]{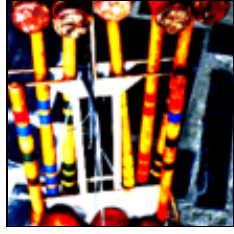}}
  \subfigure{\includegraphics[width=0.12\textwidth]{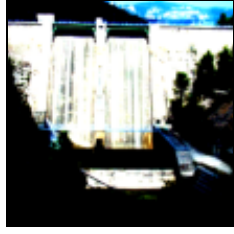}} 
  \subfigure{\includegraphics[width=0.12\textwidth]{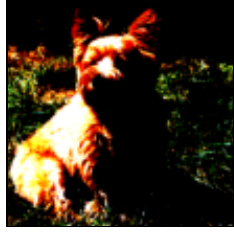}} \\ 
  \subfigure{\includegraphics[width=0.12\textwidth]{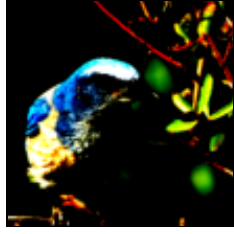}}
  \subfigure{\includegraphics[width=0.12\textwidth]{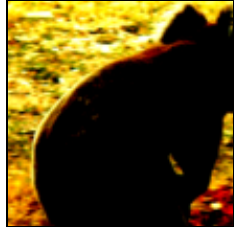}}
  \subfigure{\includegraphics[width=0.12\textwidth]{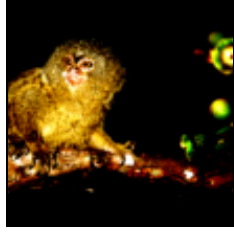}} 
  \subfigure{\includegraphics[width=0.12\textwidth]{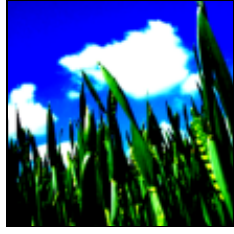}} 
  \subfigure{\includegraphics[width=0.12\textwidth]{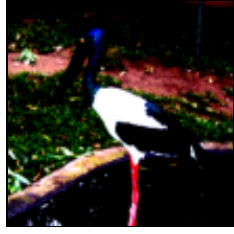}}
  \subfigure{\includegraphics[width=0.12\textwidth]{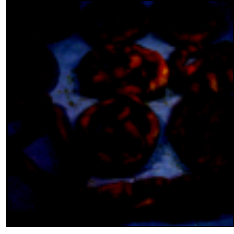}} 
  \subfigure{\includegraphics[width=0.12\textwidth]{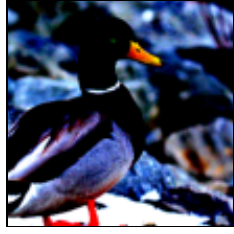}} \\ 
  \subfigure{\includegraphics[width=0.12\textwidth]{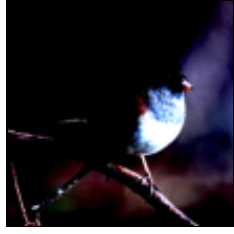}}
  \subfigure{\includegraphics[width=0.12\textwidth]{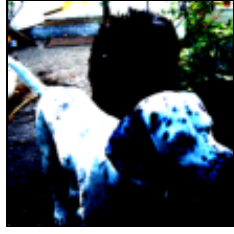}}
  \subfigure{\includegraphics[width=0.12\textwidth]{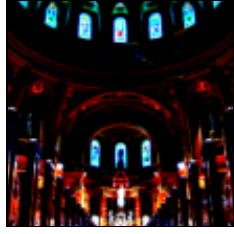}} 
  \subfigure{\includegraphics[width=0.12\textwidth]{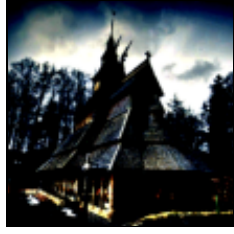}} 
  \subfigure{\includegraphics[width=0.12\textwidth]{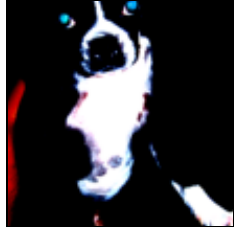}}
  \subfigure{\includegraphics[width=0.12\textwidth]{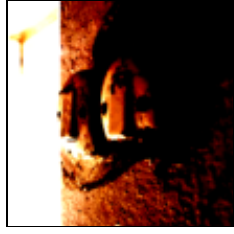}} 
  \subfigure{\includegraphics[width=0.12\textwidth]{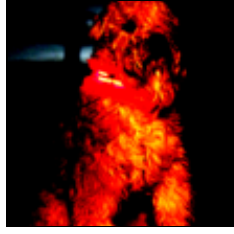}} \\ 
  \subfigure{\includegraphics[width=0.12\textwidth]{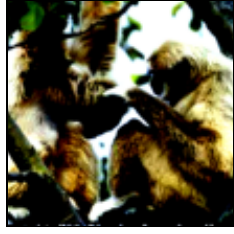}}
  \subfigure{\includegraphics[width=0.12\textwidth]{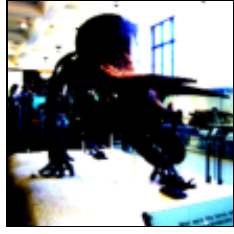}}
  \subfigure{\includegraphics[width=0.12\textwidth]{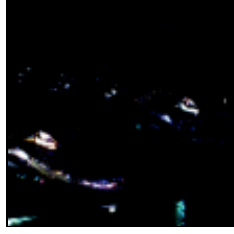}} 
  \subfigure{\includegraphics[width=0.12\textwidth]{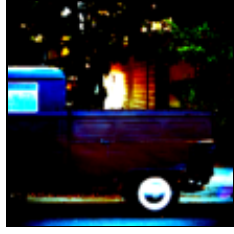}} 
  \subfigure{\includegraphics[width=0.12\textwidth]{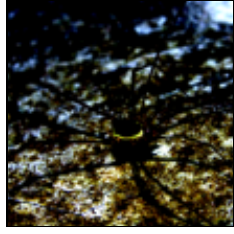}}
  \subfigure{\includegraphics[width=0.12\textwidth]{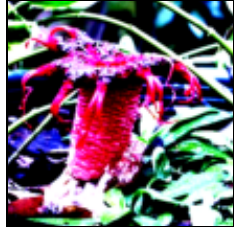}} 
  \subfigure{\includegraphics[width=0.12\textwidth]{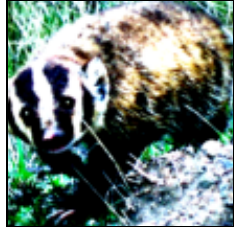}} \\ 
  \subfigure{\includegraphics[width=0.12\textwidth]{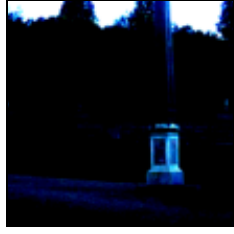}}
  \subfigure{\includegraphics[width=0.12\textwidth]{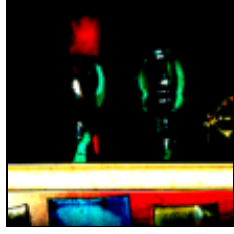}}
  \subfigure{\includegraphics[width=0.12\textwidth]{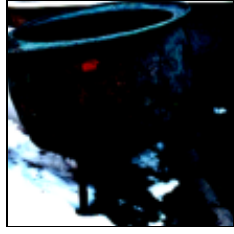}} 
  \subfigure{\includegraphics[width=0.12\textwidth]{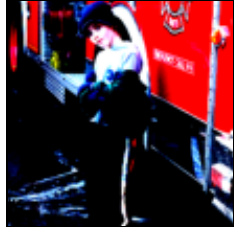}} 
  \subfigure{\includegraphics[width=0.12\textwidth]{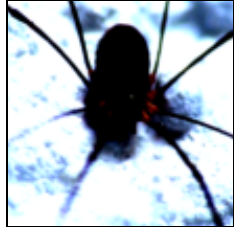}}
  \subfigure{\includegraphics[width=0.12\textwidth]{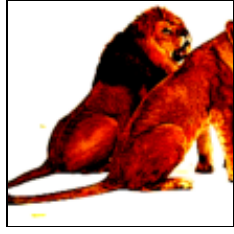}} 
  \subfigure{\includegraphics[width=0.12\textwidth]{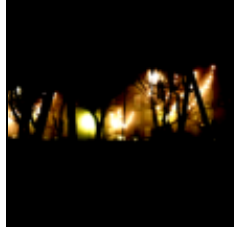}} \\ 
  \subfigure{\includegraphics[width=0.12\textwidth]{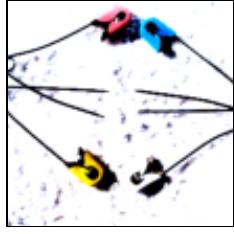}}
  \subfigure{\includegraphics[width=0.12\textwidth]{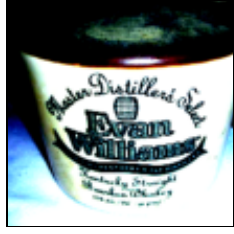}}
  \subfigure{\includegraphics[width=0.12\textwidth]{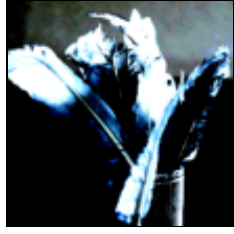}} 
  \subfigure{\includegraphics[width=0.12\textwidth]{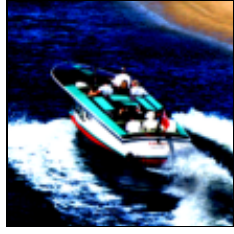}} 
  \subfigure{\includegraphics[width=0.12\textwidth]{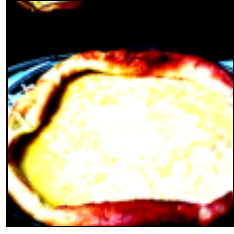}}
  \subfigure{\includegraphics[width=0.12\textwidth]{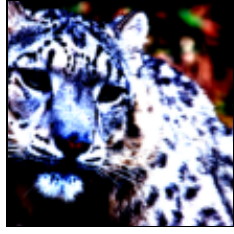}} 
  \subfigure{\includegraphics[width=0.12\textwidth]{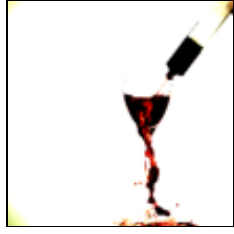}}
  \caption{More adversarial examples constructed by the proposed method for ImageNet, with $\epsilon = 4/255$. }
\end{figure}


\end{document}
